\documentclass[10pt,twocolumn,letterpaper]{article}

\usepackage{wacv}              

\usepackage{graphicx}
\usepackage{amsmath}
\usepackage{amssymb}
\usepackage{booktabs}

\usepackage{colortbl}
\usepackage{multirow}
\usepackage{adjustbox, makecell}
\usepackage{pifont}
\usepackage{array}
\usepackage{xcolor}
\usepackage{wrapfig}
\usepackage{microtype}      
\usepackage{subcaption}
\usepackage[font=small,labelfont=bf]{caption}
\usepackage{placeins}
\usepackage[accsupp]{axessibility}

\definecolor{colorexcellent}{HTML}{6aa84f} 
\definecolor{colorgood}{HTML}{b6d7a8}      
\definecolor{colorreasonable}{HTML}{ffe599} 
\definecolor{colorfair}{HTML}{ea9999}      
\definecolor{colorpoor}{HTML}{e06666}      

\definecolor{wacvblue}{rgb}{0.21,0.49,0.74}
\usepackage[pagebackref,breaklinks,colorlinks,allcolors=wacvblue]{hyperref}

\def\wacvPaperID{2850} 
\def\confName{WACV}
\def\confYear{2026}

\title{TalkingHeadBench:  A Multi-Modal Benchmark \& Analysis of Talking-Head DeepFake Detection}

\author{
\textbf{Xinqi Xiong$^{1}$*†}, 
\textbf{Prakrut Patel$^{1}$*}, 
\textbf{Qingyuan Fan$^{1}$*}, 
\textbf{Amisha Wadhwa$^{1}$*}, 
\textbf{Sarathy Selvam$^{1}$}, \\
\textbf{Xiao Guo$^{2}$}, 
\textbf{Luchao Qi$^{1}$}, 
\textbf{Xiaoming Liu$^{2}$}, 
\textbf{Roni Sengupta$^{1}$†} \\
\\[-3mm]
$^{1}$University of North Carolina at Chapel Hill \quad
$^{2}$Michigan State University \\
\tt\small \{xxiong, prakrut, lqi, ronisen\}@cs.unc.edu \\
\tt\small \{qfan, wamish, sarathy\}@unc.edu \quad
\tt\small \{guoxia11, liuxm\}@msu.edu \\
\\[-6mm]
\tt\small *Equal contribution. \quad †Corresponding Authors.
}

\begin{document}
\maketitle
\begin{abstract}

The rapid advancement of talking-head deepfake generation fueled by advanced generative models has elevated the realism of synthetic videos to a level that poses substantial risks in domains such as media, politics, and finance. However, current benchmarks for deepfake talking-head detection fail to reflect this progress, relying on outdated generators and offering limited insight into model robustness and generalization. 
We introduce TalkingHeadBench, a new benchmark designed to address this gap, featuring talking-head videos from six modern generators, with an additional two emerging generators used exclusively for testing generalization. The dataset is built on an expert-led curation process that filters over 60\% of samples to remove videos with noticeable artifacts, presenting a more difficult challenge for detectors. Our evaluation protocols are designed to measure generalization across identity and generator shifts. Benchmarking seven state-of-the-art detectors reveals that models with high accuracy on older datasets like FaceForensics++ show a significant performance drop on our curated data, particularly at strict false positive rates (e.g., TPR@FPR=0.1\%). In addition, we identify a trend where detectors focus on background cues instead of facial features using Grad-CAM visualization. 
Our benchmark aims to accelerate research towards more robust and generalizable detection models in the face of rapidly evolving generative techniques.
We release our benchmark and dataset with all data splits and protocols at
\href{https://anaxqx.github.io/talkingheadbench.github.io}{https://anaxqx.github.io/talkingheadbench.github.io}.
\end{abstract}
\section{Introduction}

\begin{figure*}[!t]
\centering
\includegraphics[width=0.8\linewidth]{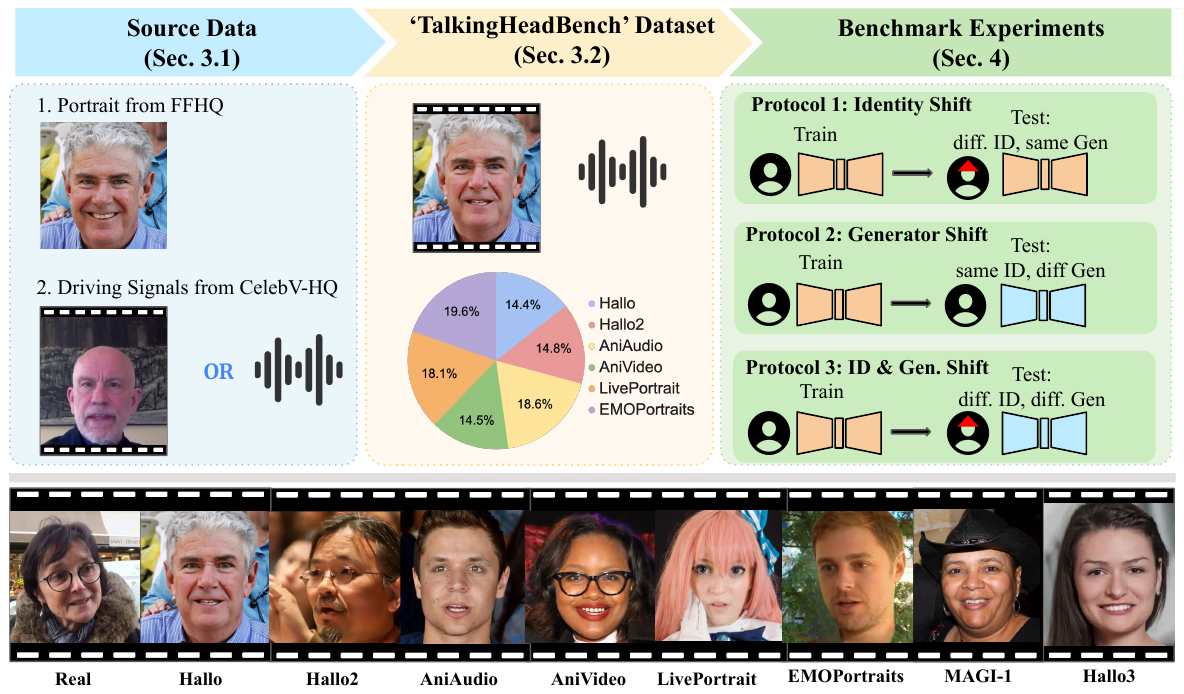}
\caption{\small Overview of the TalkingHeadBench dataset creation pipeline. We use portrait images from FFHQ~\cite{karras2019style} as source and audio or video from CelebV-HQ~\cite{zhu2022celebv} as driving signal to generate talking-head deepfake videos using 6 open-source generators for our core dataset. The images at the bottom show sample outputs from these six generators as well as from two additional emerging generators (MAGI-1~\cite{magi1} and Hallo3~\cite{cui2024hallo3}) used for evaluating generalization. We then perform multi-stage data curation to obtain 2994 high-quality deepfake videos along with 2312 real videos. Finally, we design three evaluation protocols to assess detector robustness under distribution shifts in identity and generator characteristics between training and testing sets.}
\label{fig:dataset_pipeline}
\end{figure*}

Deepfake technology, known for its ability to manipulate facial features and produce highly realistic synthetic videos, has seen rapid adoption in entertainment~\cite{liu_deepfacelab_2023, abdolahnejad_deep_2020}, marketing~\cite{kaur_deepfake_2024}, and personalized content creation~\cite{qi_mytimemachine_2024, qi_my3dgen_2025, wei_personalized_2025}. Yet, these advancements also introduce serious societal risks. Among deepfake types, talking-head deepfakes, which generate highly realistic head-and-shoulder videos of individuals speaking, are especially concerning. Driven by advanced generative models, these videos are potent tools for spreading misinformation, manipulating public discourse, and executing social engineering attacks~\cite{gambin_deepfakes_2024, saini_deepfake_2024, saif_fake_2024}. Their realism and ease of distribution make them particularly dangerous in sensitive domains like politics, finance, and media. In one real-world case, a finance worker was tricked into transferring \$25 million during a video call with a deepfake impersonating the company’s CFO~\cite{magramo_finance_2024}. These growing threats highlight the urgent need to understand talking-head deepfakes and to develop robust detection and mitigation strategies~\cite{gambin_deepfakes_2024, sirra_csslno_2024, liu_evolving_2025}.

In response to the growing threats posed by talking-head deepfakes, numerous facial deepfake benchmarks have been proposed (see Tab. \ref{tab:datasets}), which played a central role in advancing detection methods. However, most existing benchmarks are not well-suited for the modern talking-head landscape. Most deepfake datasets primarily feature face-swapping techniques \cite{rossler2019faceforensics++} rather than talking-head videos, which are generated using different techniques. Although large-scale and collected in the wild, these datasets often contain a high proportion of low-quality samples with noticeable artifacts \cite{chandra2025deepfake,yan2024df40}, making it easier to train detectors that ultimately perform poorly on more realistic deepfakes, which pose a greater threat.
This leaves a gap in reliably measuring the performance of detectors against the more realistic manipulations produced by the recent talking-head generators. 
To address these gaps, we introduce TalkingHeadBench—the first manually curated, multi-generator, multi-modal benchmark designed specifically for evaluating detector robustness against state-of-the-art talking-head deepfakes.

Our benchmark features high-quality synthetic videos generated with six academic talking-head generators using both audio- and video-driven diffusion and transformer-based models~\cite{drobyshev2024emoportraits, cui2024hallo2, xu2024hallo, wei2024aniportrait, guo2024liveportrait} in the training dataset. To further test generalization, we evaluate models against two emerging generators: an additional academic model (Hallo3~\cite{cui2024hallo3}) and a commercial one (MAGI-1~\cite{magi1}). To ensure dataset quality, we use a five-stage expert curation protocol that removes over 60\% of generated videos based on pre-defined artifact criteria. The result is a challenging dataset of 2,994 high-quality talking-head deepfakes from eight academic and commercial generators. This emphasis on quality over quantity creates a more realistic evaluation setting.
We also carefully design evaluation protocols to assess the generalization and robustness of state-of-the-art (SOTA) deepfake detectors~\cite{dong2023implicit, shao2025deepfake, xu2023tall, xu2024learning, liu2024lips} across identity and generator shifts in training and testing data.
In addition to benchmarking, we present a detailed error analysis using Grad-CAM~\cite{Selvaraju_2019} to identify common failure cases of SOTA detectors on different generators and inform future improvements in detection strategies.
We then note the limitations of SOTA detectors and identify areas of improvement, and provide guidelines on what performance future detectors should aspire to and how to analyze them.

We open-source the TalkingHeadBench benchmark with train-test data for all generators across multiple protocols on our project page.
In the near future, we aim to build a more active benchmark repository by a) developing API/code/metrics that also allow authors of new generators to evaluate the `detectability' of their videos;
b) setting up two leaderboards comparing generator and detector performance;
c) updating the list of generators and detectors every 6 months to accelerate the development cycle.

Our main contributions are:
\begin{itemize}[noitemsep,topsep=0pt]
    \item \textbf{A highly curated benchmark for modern talking-head generation.} We release TalkingHeadBench, a multi-modal and multi-generator dataset of 2,994 high-quality deepfake videos, filtered from a much larger pool of deepfake videos using a multi-stage expert-led curation process.
    
    \item \textbf{An analysis of detector performance at strict thresholds.} We show that standard metrics like AUC can obscure poor performance. On our benchmark, SOTA detectors exhibit sharp performance drops at more practical, low-FPR thresholds (e.g., TPR@FPR=0.1\%).

    \item \textbf{A study of generalization across identity and generator shifts.} Our structured protocols quantify the performance drop of detectors across these distribution shifts, showing the combined effect remains a difficult challenge.

    \item \textbf{An explainability analysis identifying semantic drift.} Using Grad-CAM, we show that as generator quality improves, detectors tend to shift their focus from facial regions to background artifacts, exposing a common failure.
\end{itemize}
In summary, our benchmark demonstrates that while state-of-the-art detectors attain high accuracy on existing deepfake detection benchmarks \cite{rossler2019faceforensics++}, their performance drops markedly on TalkingHeadBench. This degradation stems from the inclusion of advanced multi-modal generators, deliberate shifts in identity and generator between training and testing splits, and careful curation that eliminates obvious artifacts in the deepfakes. Together, these factors highlight the need for developing more robust deepfake detectors and establish our benchmark as a foundation for driving future research in this direction.

\vspace{-0.2em}
\begin{table*}[t!]
\centering
\caption{\small Overview of Facial Deepfake Benchmarks. TalkingHeadBench is the first manually curated, multi-modal, multi-generator benchmark focused on talking-head deepfakes, specifically designed to evaluate detector robustness and generalization. Curation removes $\sim$60\% of low-quality samples with obvious artifacts, ensuring a more realistic and challenging benchmark. Generators marked with \ding{51}* use lip-sync-only techniques. }
\label{tab:datasets}
 \resizebox{0.85\linewidth}{!}
{%
\begin{tabular}{c|c|c|c|c|c|c|c}
\toprule
Dataset & Talking-head & \# of Generators & Modality &  Curation & Year &  Real \# &  Fake \# \\
\midrule
DeepFake-eval-2024~\cite{chandra2025deepfake} &
\ding{55} &N/A&Video/Audio/Image &\ding{51}&2025 & 1,208 & 767\\

FakeAVCeleb~\cite{khalid2021fakeavceleb} & \ding{55} & Multiple & Audio/Video & \ding{51} & 2021 & 500 & 19,500\\
FaceForensics++~\cite{rossler2019faceforensics++} & \ding{55} & Multiple & Video & \ding{51} & 2019 & 1,000 & 4,000\\
LAV-DF~\cite{cai2022you} &\ding{51}* & Single & Audio/Video & \ding{55} & 2022 & 36,431 & 99,873\\

AV-Deepfake1M~\cite{cai2024av} &\ding{51}* & Single & Audio/Video & \ding{55}& 2023 & 286,721 & 860,039\\

PolyGlotFake~\cite{hou2024polyglotfakenovelmultilingualmultimodal} & \ding{51}* & Multiple & Audio & \ding{55}& 2024 & 766 & 14,472\\

DF-40 ~\cite{yan2024df40} &
\ding{51}
& Multiple & Video/Image &\ding{55} & 2024 & 1,590 & 1M+ \\
\rowcolor{lightgray}
TalkingHeadBench (Ours) & \ding{51} & Multiple & Audio/Video & \ding{51} & 2025 & 2,312 & 2,994\\

\bottomrule
\end{tabular}%
}
\vspace{-3mm}
\end{table*}
\section{Related Work}

Deepfake generation has leveraged various facial manipulation techniques over the years. 
Among them, face swapping methods~\cite{nirkin2019fsgan,rosberg2023facedancer,chen2020simswap} are widely used, combining facial regions of various sources to create synthetic videos. 
They often introduce visible artifacts such as boundary inconsistency, lighting mismatch, or texture irregularity, which modern detectors can learn to exploit easily.

Talking-head generation presents a different set of challenges. Unlike face-swapping, which inserts a swapped facial region and often leaves detectable boundary artifacts, talking-head models synthesize the entire face in motion. These manipulations are driven by audio or video signals and are characterized by their temporal properties and speech-synchronized movements. This makes them a distinct class of forgery that requires evaluation against detectors trained to spot not just spatial inconsistencies, but also subtle temporal and behavioral anomalies. Our benchmark is designed specifically to address this category of deepfake.

Recently, talking-head generation has emerged as a more sophisticated approach. These models synthesize the full face, including expressions and lip motion, from a static portrait using audio or video driving signals by eliminating the obvious seam-based flaws of earlier methods. State-of-the-art diffusion-based talking-head models (see Tab.~\ref{tab:generators}) produce highly realistic outputs with synchronized audio-visual features, significantly raising the detection difficulty.

Despite these advances, existing deepfake benchmarks remain limited in scope. As shown in Tab.~\ref{tab:datasets}, most prior datasets either focus on older GAN-based generators or target narrow manipulation types such as face swapping ({\it e.g.}, FakeAVCeleb~\cite{khalid2021fakeavceleb} and FaceForensics++~\cite{rossler2019faceforensics++}) or scrape social media content without controlling of generators ({\it e.g.}, DeepFake-eval-2024~\cite{chandra2025deepfake}). While newer datasets like LAV-DF~\cite{cai2023glitch}, AV-Deepfake1M~\cite{cai2024av}, and PolyGlobFake~\cite{hou2024polyglotfakenovelmultilingualmultimodal} incorporate talking-head deepfakes, they mostly emphasize lip-sync-driven synthesis and lack broader modality or generator diversity. Further, these datasets are often massive but uncurated, resulting in many low-quality samples. In contrast, our dataset is built using a curated collection of cutting-edge diffusion-based talking-head generators spanning both audio and video-driven modalities. Through manual inspection, we discarded over 60\% of generated videos due to visible artifacts (see Tab. \ref{tab:dataset_stats}), ensuring that only high-quality, realistic samples remain—thereby offering a more meaningful and challenging benchmark.

Recent detection methods (see Tab.~\ref{tab:detectors}), though  effective on static and unimodal datasets like FaceForensics++~\cite{rossler2019faceforensics++}, have not been thoroughly evaluated against multi-modal diffusion-based deepfakes. To fill this gap, we introduce a comprehensive benchmark that reflects the current generation landscape. Our results show that SOTA detectors, which achieve near-perfect accuracy on traditional benchmarks like FaceForensics++, struggle significantly when tested on our dataset, highlighting poor robustness to identity and generator shifts and calling for stronger generalization capabilities.

\begin{table}[t!]
\small
\centering
\caption{\small Overview of talking-head generators, \colorbox{lightgray}{highlighted} used for creating the train and test sets of TalkingHeadBench.}
\label{tab:generators}
 \resizebox{\columnwidth}{!}
{%
\begin{tabular}{c|c|l|c|c}
\toprule
Modality& Framework & Generator & Venue & Code Availability \\
\midrule
\rowcolor{lightgray}
Image, Audio & Diffusion & Hallo~\cite{xu2024hallo} & arXiv24 & \ding{51}\\ 

\rowcolor{lightgray}
Image, Audio & Diffusion & Hallo2~\cite{cui2024hallo2} & ICLR25 & \ding{51}\\

\rowcolor{lightgray}
Image, Audio, Text  & Diffusion & Hallo3~\cite{cui2024hallo3} &  CVPR25 & \ding{51}\\
Image, Video  & Diffusion & X-Portrait~\cite{xie2024x} &  SIGGRAPH24 & \ding{51}  \\

\rowcolor{lightgray}
Image, Video  & Non-diffusion & LivePortrait~\cite{guo2024liveportrait} & arXiv24 & \ding{51}
\\
\rowcolor{lightgray}
Image, Video  & Non-diffusion & EMOPortraits~\cite{drobyshev2024emoportraits} & CVPR24 & \ding{51}\\

 Image, Video  & Non-diffusion & MCNet~\cite{hong2023implicit} &  ICCV23 & \ding{51}  \\
 Image, Video & Non-diffusion & FSRT~\cite{rochow2024fsrt} & CVPR24 & \ding{55}\\

\rowcolor{lightgray} 
Image, Audio/Video & Diffusion & AniPortrait~\cite{wei2024aniportrait} & arXiv24 & \ding{51}\\

Image, Audio/Video & Diffusion & SkyReels-A1~\cite{qiu2025skyreelsa1expressiveportraitanimation} & arXiv25 & \ding{55}\\
\rowcolor{lightgray} 
Image, Text &Diffusion&MAGI-1~\cite{magi1}&N/A& \ding{51}\\
\bottomrule
\end{tabular}%
}
\\
\end{table}
\begin{table}[t!]
\centering
\normalsize
\caption{\small Overview of talking-head deepfake detectors, \colorbox{lightgray}{highlighted} were used in the paper.}
\label{tab:detectors}
\resizebox{\columnwidth}{!}{%
\begin{tabular}{c|c|l|c|c}
\toprule
Modality& Framework & Detector & Venue & Code Availability \\
\midrule
Image & CNN & SBI~\cite{shiohara2022detecting} & CVPR22 & \ding{51}\\
\rowcolor{lightgray}
Image & Neural Network & CADDM~\cite{dong2023implicit} & CVPR23 & \ding{51}\\
\rowcolor{lightgray}
Image & Neural Network & HiFiNet\cite{hifi_net_xiaoguo}//HiFiNet++\cite{xiao_hifinet_plusplus} & CVPR23/IJCV24 & \ding{51}\\
\rowcolor{lightgray}
Image & Transformer & DeepFake-Adapter~\cite{shao2025deepfake} & IJCV25 & \ding{51}\\

Image & Transformer& GenConVit~\cite{deressa2025genconvitdeepfakevideodetection} & arXiv25& \ding{51}\\
Image &Transformer &RFFR~\cite{shi2025real}& PR25 &  \ding{51}\\
Image & Neural network & LAA-Net~\cite{nguyen2024laa} & CVPR24 & \ding{51}\\
\rowcolor{lightgray}
Video  & Transformer & TALL~\cite{xu2023tall}/TALL++~\cite{xu2024learning} &  ICCV23/IJCV24 & \ding{51}\\
\rowcolor{lightgray}
Video & CNN & AltFreezing~\cite{wang2023altfreezing} & CVPR23 &  \ding{55} \\
Video & CNN/Transformer & NACO~\cite{zhang2024learning} & ECCV24 & \ding{55}\\
\rowcolor{lightgray}
Video & Neural network & MM-Det~\cite{song2025learningmultimodalforgeryrepresentation} &  NeurIPS24 & \ding{51} \\

Video & Neural network & StyleFlow~\cite{choi2024exploiting} & CVPR24 & \ding{55}\\

\rowcolor{lightgray}
Audio/Video & Transformer & LipFD~\cite{liu2024lips} & NeurIPS 2024 & \ding{51}\\
\bottomrule
\end{tabular}%
}
\end{table}

\section{Dataset Creation}
\label{sec:dataset}



This section details the creation process of our dataset, including data sources, generation methods, curation procedures, statistics, and data splits.
See Fig.~\ref{fig:dataset_pipeline} for the overview.

\subsection{Source Data and Generation Overview}
\label{subsec:source_data}
To create a diverse deepfake benchmark dataset, we utilize publicly available datasets containing images from FFHQ (Flickr-Faces-HQ)~\cite{karras2019style} and videos of real people and audio of people talking from CelebV-HQ~\cite{zhu2022celebv}, processed by a diverse suite of contemporary deepfake generators. 
We select six open-source generators: Hallo~\cite{xu2024hallo}, Hallo2~\cite{cui2024hallo2}, AniPortrait (Audio-driven) (AniAudio)~\cite{wei2024aniportrait}, AniPortrait (Video-driven) (AniVideo)~\cite{wei2024aniportrait}, LivePortrait~\cite{guo2024liveportrait}, and EMOPortraits~\cite{drobyshev2024emoportraits}, aiming for diversity in both diffusion-based and non-diffusion-based frameworks and driving signal modalities in both video and audio, as summarized in Tab.~\ref{tab:generators}. 
We choose these generators because they are the latest SoTA generators with publicly available code covering varying modalities and frameworks.

Additionally, we test generalization against two entirely unseen models: Hallo3~\cite{cui2024hallo3}, a recent academic generator, and MAGI-1~\cite{magi1}, a commercial text-to-video model. These are used exclusively in our test set to evaluate out-of-distribution robustness. Owing to computational and cost limitations, these test sets are relatively small in size, but they nevertheless provide a meaningful probe of model generalization. We also explore other generators such as X-Portrait~\cite{xie2024x} and MCNet~\cite{hong2023implicit}, but do not include them in our dataset due to significant artifacts observed.
\vspace{-2mm}

\begin{table}[ht!]  
\centering
\caption{\small TalkingHeadBench dataset statistics across generators.}
\label{tab:dataset_stats}

\begingroup
\small
\setlength{\tabcolsep}{4pt}      
\renewcommand{\arraystretch}{1.05}

\adjustbox{max width=\columnwidth}{%
\begin{tabular}{@{}lcccc@{}}
\toprule
\multicolumn{1}{c}{Data Type} & \multicolumn{1}{c}{\# Total} & \multicolumn{1}{c}{\# Train} & \multicolumn{1}{c}{\# Test} & \multicolumn{1}{c}{Ratio Removed} \\
\midrule
Hallo~\cite{xu2024hallo}                      & 420  & 303 & 117 & $\sim42\%$ \\
Hallo2~\cite{cui2024hallo2}                   & 432  & 327 & 105 & $\sim60\%$ \\
AniPortrait (Audio)~\cite{wei2024aniportrait} & 542  & 396 & 146 & $\sim66.7\%$ \\
AniPortrait (Video)~\cite{wei2024aniportrait} & 422  & 314 & 108 & $\sim80\%$ \\
LivePortrait~\cite{guo2024liveportrait}      & 529  & 352 & 177 & $\sim60\%$ \\
EMOPortraits~\cite{drobyshev2024emoportraits} & 573  & 381 & 192 & $\sim50\%$ \\
\midrule
MAGI-1~\cite{magi1}                           & 66   & 0   & 66  & $\sim18\%$ \\
Hallo3~\cite{cui2024hallo3}                           & 10   & 0   & 10  & $\sim10\%$ \\
\midrule
\textbf{TOTAL Fakes}                          & 2994 & 2073& 921 & $\sim63\%$ \\
\bottomrule
\end{tabular}%
}
\endgroup
\vspace{-3ex}
\end{table}
\vspace{0.7em}
\subsection{Data Curation and Final Dataset Statistics}
\label{subsec:generation}
\vspace{0.7em}

In order to create a high-quality dataset, we used a five-stage curation pipeline conducted by experts on our team. This process removed approximately 63\% of the nearly 8,000 videos we initially generated, ensuring the final dataset is not easily solved by detectors that rely on simple artifacts.

\begin{enumerate}[noitemsep,topsep=0pt]
 
\item \textbf{Generator-Specific Artifact Definition:} We analyzed videos from each generator to create a taxonomy of common artifacts, such as severe background warping or unnatural static hair. This formed a consistent set of rejection criteria, which are detailed in the supplementary.

\item \textbf{Primary Review Pass:} A curator reviewed all videos from a specific generator, removing samples that matched the artifact definitions or had other global distortions.

\item \textbf{Cross-Validation by Experts:} To ensure consistency, two other curators independently reviewed the remaining videos and removed any additional low-quality samples. Disagreements were resolved through discussion.

\item \textbf{Detector-Based Final Checkpoint:} To further refine the dataset's difficulty, we passed the remaining videos through four pre-trained SOTA detectors. Any video detected with high confidence ($\geq$ 90\% fake probability) across all four detectors was removed. This step filtered out samples that, while appearing plausible to humans, contained subtle but easily machine-detectable cues.

\item \textbf{Identity-Separated Splits}: The final curated set was partitioned into training and testing splits, with strict separation enforced by disallowing any overlap in source image-driving signal pairings across splits. The final dataset statistics are detailed in Tab.~\ref{tab:dataset_stats}.
\end{enumerate}

\section{Benchmark Experiments}
\label{sec:experiments}

\subsection{Evaluation Setup}

\begin{table*}[t]
\centering
\caption{\textbf{Overall benchmark.} Three protocols reported as stacked subtables. Best numbers per generator in \textbf{bold}.}
\label{tab:mega_all_protocols}
\setlength{\tabcolsep}{3.5pt}
\renewcommand{\arraystretch}{1.08}
\scriptsize

\begin{subtable}{\textwidth}
\centering
\caption{\textbf{Protocol 1} — Identity Shift Only.}
\label{tab:m1_colored_updated}
\adjustbox{max width=\textwidth}{%
\begin{tabular}{l cccc cccc cccc cccc cccc cccc}
\toprule
& \multicolumn{4}{c}{AniAudio~\cite{wei2024aniportrait}}
& \multicolumn{4}{c}{AniVideo~\cite{wei2024aniportrait}}
& \multicolumn{4}{c}{Hallo~\cite{xu2024hallo}}
& \multicolumn{4}{c}{Hallo2~\cite{cui2024hallo2}}
& \multicolumn{4}{c}{EMOPortraits~\cite{drobyshev2024emoportraits}}
& \multicolumn{4}{c}{LivePortrait~\cite{guo2024liveportrait}} \\
\cmidrule(lr){2-5}\cmidrule(lr){6-9}\cmidrule(lr){10-13}\cmidrule(lr){14-17}\cmidrule(lr){18-21}\cmidrule(lr){22-25}
Detector & AUC & T1 & T0.1 & Brier
         & AUC & T1 & T0.1 & Brier
         & AUC & T1 & T0.1 & Brier
         & AUC & T1 & T0.1 & Brier
         & AUC & T1 & T0.1 & Brier
         & AUC & T1 & T0.1 & Brier \\
\midrule
CADDM~\cite{dong2023implicit}
& 0.90 & \cellcolor{colorpoor}{0.55} & \cellcolor{colorpoor}{0.25} & 0.17
& 0.83 & \cellcolor{colorpoor}{0.39} & \cellcolor{colorpoor}{0.39} & 0.21
& 0.94 & \cellcolor{colorpoor}{0.65} & \cellcolor{colorpoor}{0.64} & 0.16
& 0.96 & \cellcolor{colorpoor}{0.45} & \cellcolor{colorpoor}{0.45} & 0.11
& 0.94 & \cellcolor{colorpoor}{0.27} & \cellcolor{colorpoor}{0.23} & 0.17
& 0.95 & \cellcolor{colorpoor}{0.59} & \cellcolor{colorpoor}{0.42} & 0.13 \\
HiFi-Net~\cite{hifi_net_xiaoguo} & 0.95 & \cellcolor{colorpoor}{0.47} & \cellcolor{colorpoor}{0.27} & 0.09 & 0.95  & \cellcolor{colorpoor}{0.66} & \cellcolor{colorpoor}{0.41} & 0.09 & 0.93 & \cellcolor{colorpoor}{0.39} & \cellcolor{colorpoor}{0.19} & 0.11 & 0.92 & \cellcolor{colorpoor}{0.22} & \cellcolor{colorpoor}{0.12} & 0.12 & 0.81 & \cellcolor{colorpoor}{0.03} & \cellcolor{colorpoor}{0.01} & 0.25 & 0.90 & \cellcolor{colorpoor}{0.26} & \cellcolor{colorpoor}{0.04} & 0.12 \\
AltFreezing~\cite{wang2023altfreezing} &1.00 & \cellcolor{colorreasonable}{0.93} & \cellcolor{colorpoor}{0.71} & 0.03 & 1.00 & \cellcolor{colorreasonable}{0.91} & \cellcolor{colorfair}{0.75} & 0.05 & 1.00 & \cellcolor{colorgood}{0.97} & \cellcolor{colorgood}{0.97} & 0.03 & 1.00 & \cellcolor{colorgood}{0.95} & \cellcolor{colorreasonable}{0.88} & 0.05 & 1.00 & \cellcolor{colorreasonable}{0.93} & \cellcolor{colorreasonable}{0.89} & 0.04 & 1.00 & \cellcolor{colorreasonable}{0.94} & \cellcolor{colorreasonable}{0.89} & 0.04 \\
TALL~\cite{xu2023tall}
& \textbf{1.00} & \cellcolor{colorexcellent}{\textbf{1.00}} & \cellcolor{colorexcellent}{\textbf{1.00}} & 0.01
& \textbf{1.00} & \cellcolor{colorexcellent}{\textbf{1.00}} & \cellcolor{colorexcellent}{\textbf{1.00}} & 0.01
& \textbf{1.00} & \cellcolor{colorexcellent}{\textbf{0.99}} & \cellcolor{colorexcellent}{\textbf{0.99}} & 0.01
& \textbf{1.00} & \cellcolor{colorexcellent}{\textbf{0.99}} & \cellcolor{colorgood}{\textbf{0.98}} & 0.01
& \textbf{1.00} & \cellcolor{colorgood}{0.98} & \cellcolor{colorgood}{\textbf{0.98}} & 0.05
& \textbf{1.00} & \cellcolor{colorgood}{0.97} & \cellcolor{colorgood}{\textbf{0.96}} & 0.01 \\
LipFD~\cite{liu2024lips}
& 0.98 & \cellcolor{colorfair}{0.75} & \cellcolor{colorpoor}{0.58} & 0.05
& 0.99 & \cellcolor{colorpoor}{0.74} & \cellcolor{colorpoor}{0.38} & 0.05
& 0.98 & \cellcolor{colorpoor}{0.72} & \cellcolor{colorpoor}{0.54} & 0.05
& 0.99 & \cellcolor{colorfair}{0.83} & \cellcolor{colorpoor}{0.62} & 0.04
& 0.99 & \cellcolor{colorfair}{0.78} & \cellcolor{colorpoor}{0.50} & 0.05
& 0.98 & \cellcolor{colorfair}{0.84} & \cellcolor{colorpoor}{0.56} & 0.05 \\
DF-Adapter~\cite{shao2025deepfake}
& 0.99 & \cellcolor{colorpoor}{0.58} & \cellcolor{colorpoor}{0.06} & 0.01
& \textbf{1.00} & \cellcolor{colorexcellent}{\textbf{1.00}} & \cellcolor{colorpoor}{0.15} & 0.01
& \textbf{1.00} & \cellcolor{colorexcellent}{\textbf{0.99}} & \cellcolor{colorpoor}{0.11} & 0.01
& \textbf{1.00} & \cellcolor{colorexcellent}{\textbf{0.99}} & \cellcolor{colorgood}{\textbf{0.98}} & 0.01
& \textbf{1.00} & \cellcolor{colorexcellent}{\textbf{0.99}} & \cellcolor{colorpoor}{0.25} & 0.01
& \textbf{1.00} & \cellcolor{colorgood}{\textbf{0.98}} & \cellcolor{colorpoor}{0.41} & 0.01 \\
MM-Det~\cite{song2025learningmultimodalforgeryrepresentation} & 0.97 & \cellcolor{colorpoor}{0.65} & \cellcolor{colorpoor}{0.26} & 0.14  & 0.97 & \cellcolor{colorpoor}{0.55} & \cellcolor{colorpoor}{0.19} & 0.15  & 0.97 & \cellcolor{colorpoor}{0.63} & \cellcolor{colorpoor}{0.26} & 0.14  & 0.96 & \cellcolor{colorpoor}{0.57} & \cellcolor{colorpoor}{0.21} & 0.15  & 0.93 & \cellcolor{colorpoor}{0.34} & \cellcolor{colorpoor}{0.11} & 0.14 & 0.91 & \cellcolor{colorpoor}{0.28} & \cellcolor{colorpoor}{0.04} & 0.15 \\
\bottomrule
\end{tabular}}
\end{subtable}

\vspace{3pt}

\begin{subtable}{\textwidth}
\centering
\caption{\textbf{Protocol 2} — Generator Shift Only.}
\label{tab:m2_scheme2_T1_colored}
\adjustbox{max width=\textwidth}{%
\begin{tabular}{l cccc cccc cccc cccc cccc cccc}
\toprule
& \multicolumn{4}{c}{AniAudio~\cite{wei2024aniportrait}}
& \multicolumn{4}{c}{AniVideo~\cite{wei2024aniportrait}}
& \multicolumn{4}{c}{Hallo~\cite{xu2024hallo}}
& \multicolumn{4}{c}{Hallo2~\cite{cui2024hallo2}}
& \multicolumn{4}{c}{EMOPortraits~\cite{drobyshev2024emoportraits}}
& \multicolumn{4}{c}{LivePortrait~\cite{guo2024liveportrait}} \\
\cmidrule(lr){2-5}\cmidrule(lr){6-9}\cmidrule(lr){10-13}\cmidrule(lr){14-17}\cmidrule(lr){18-21}\cmidrule(lr){22-25}
Detector & AUC & T1 & T0.1 & Brier
         & AUC & T1 & T0.1 & Brier
         & AUC & T1 & T0.1 & Brier
         & AUC & T1 & T0.1 & Brier
         & AUC & T1 & T0.1 & Brier
         & AUC & T1 & T0.1 & Brier \\
\midrule
CADDM~\cite{dong2023implicit} & 0.88 & \cellcolor{colorpoor}{0.32} & \cellcolor{colorpoor}{0.02} & 0.18
& 0.94 & \cellcolor{colorpoor}{0.27} & \cellcolor{colorpoor}{0.18} & 0.09
& 0.97 & \cellcolor{colorpoor}{0.61} & \cellcolor{colorpoor}{0.52} & 0.08
& 0.94 & \cellcolor{colorpoor}{0.37} & \cellcolor{colorpoor}{0.14} & 0.12
& 0.79 & \cellcolor{colorpoor}{0.02} & \cellcolor{colorpoor}{0.00} & 0.21
& 0.98 & \cellcolor{colorpoor}{0.71} & \cellcolor{colorpoor}{0.63} & 0.06 \\
HiFi-Net~\cite{hifi_net_xiaoguo} & 0.88 & \cellcolor{colorpoor}{0.08} & \cellcolor{colorpoor}{0.01} & 0.16 & 0.82 & \cellcolor{colorpoor}{0.16} & \cellcolor{colorpoor}{0.06} & 0.18  & 0.82 & \cellcolor{colorpoor}{0.02} & \cellcolor{colorpoor}{0.00} & 0.20 & 0.48 & \cellcolor{colorpoor}{0.02} & \cellcolor{colorpoor}{0.00} & 0.45 & 0.58 & \cellcolor{colorpoor}{0.01} & \cellcolor{colorpoor}{0.00} & 0.34  & 0.85 & \cellcolor{colorpoor}{0.01} & \cellcolor{colorpoor}{0.01} & 0.20 \\
AltFreezing~\cite{wang2023altfreezing} & 1.00 & \cellcolor{colorgood}{0.97} & \cellcolor{colorfair}{0.82} & 0.02 & 1.00 & \cellcolor{colorreasonable}{0.94} & \cellcolor{colorpoor}{0.66} & 0.02 & 1.00 & \cellcolor{colorgood}{0.98} & \cellcolor{colorreasonable}{0.87} & 0.01 & 0.98 & \cellcolor{colorpoor}{0.71} & \cellcolor{colorpoor}{0.53} & 0.06 & 0.96 & \cellcolor{colorpoor}{0.69} & \cellcolor{colorpoor}{0.52} & 0.08 & 1.00 & \cellcolor{colorreasonable}{0.94} & \cellcolor{colorfair}{0.78} & 0.02 \\
TALL~\cite{xu2023tall} & \textbf{1.00} & \cellcolor{colorexcellent}{\textbf{1.00}} & \cellcolor{colorgood}{\textbf{0.98}} & 0.01
& \textbf{1.00} & \cellcolor{colorexcellent}{0.99} & \cellcolor{colorgood}{0.98} & 0.02
& \textbf{1.00} & \cellcolor{colorexcellent}{\textbf{1.00}} & \cellcolor{colorexcellent}{\textbf{1.00}} & 0.00
& \textbf{0.97} & \cellcolor{colorreasonable}{\textbf{0.89}} & \cellcolor{colorfair}{\textbf{0.81}} & 0.08 & 0.97 & \cellcolor{colorpoor}{0.70} & \cellcolor{colorpoor}{0.40} & 0.15
& \textbf{1.00} & \cellcolor{colorexcellent}{\textbf{1.00}} & \cellcolor{colorgood}{0.98} & 0.01 \\
LipFD~\cite{liu2024lips} & 0.99 & \cellcolor{colorfair}{0.76} & \cellcolor{colorpoor}{0.51} & 0.05
& 0.97 & \cellcolor{colorpoor}{0.62} & \cellcolor{colorpoor}{0.30} & 0.09
& 0.98 & \cellcolor{colorpoor}{0.70} & \cellcolor{colorpoor}{0.47} & 0.08
& 0.87 & \cellcolor{colorpoor}{0.50} & \cellcolor{colorpoor}{0.32} & 0.17
& 0.89 & \cellcolor{colorpoor}{0.30} & \cellcolor{colorpoor}{0.08} & 0.23
& 0.99 & \cellcolor{colorfair}{0.81} & \cellcolor{colorpoor}{0.63} & 0.05 \\
DF-Adapter~\cite{shao2025deepfake} & \textbf{1.00} & \cellcolor{colorexcellent}{\textbf{1.00}} & \cellcolor{colorfair}{0.83} & 0.01
& \textbf{1.00} & \cellcolor{colorexcellent}{\textbf{1.00}} & \cellcolor{colorexcellent}{\textbf{1.00}} & 0.00
& \textbf{1.00} & \cellcolor{colorexcellent}{\textbf{1.00}} & \cellcolor{colorpoor}{0.73} & 0.01
& \textbf{0.97} & \cellcolor{colorfair}{0.80} & \cellcolor{colorpoor}{0.68} & 0.15
& \textbf{0.99} & \cellcolor{colorreasonable}{\textbf{0.93}} & \cellcolor{colorfair}{\textbf{0.80}} & 0.13
& \textbf{1.00} & \cellcolor{colorexcellent}{\textbf{1.00}} & \cellcolor{colorexcellent}{\textbf{0.99}} & 0.01 \\
MM-Det~\cite{song2025learningmultimodalforgeryrepresentation} & 0.93 & \cellcolor{colorpoor}{0.39} & \cellcolor{colorpoor}{0.04} & 0.12 & 0.95 & \cellcolor{colorpoor}{0.36} & \cellcolor{colorpoor}{0.14} & 0.11 & 0.94 & \cellcolor{colorpoor}{0.34} & \cellcolor{colorpoor}{0.08} & 0.15 & 0.97 & \cellcolor{colorpoor}{0.51} & \cellcolor{colorpoor}{0.25} & 0.06 & 0.89 & \cellcolor{colorpoor}{0.15} & \cellcolor{colorpoor}{0.07} & 0.06 & 0.88 & \cellcolor{colorpoor}{0.23} & \cellcolor{colorpoor}{0.11} & 0.16 \\
\bottomrule
\end{tabular}}
\end{subtable}

\vspace{3pt}

\begin{subtable}{\textwidth}
\centering
\caption{\textbf{Protocol 3} — Identity \& Generator Shift Combined.}
\label{tab:m3_scheme2_T1_colored}
\adjustbox{max width=\textwidth}{%
\begin{tabular}{l cccc cccc cccc cccc cccc cccc}
\toprule
& \multicolumn{4}{c}{AniAudio~\cite{wei2024aniportrait}}
& \multicolumn{4}{c}{AniVideo~\cite{wei2024aniportrait}}
& \multicolumn{4}{c}{Hallo~\cite{xu2024hallo}}
& \multicolumn{4}{c}{Hallo2~\cite{cui2024hallo2}}
& \multicolumn{4}{c}{EMOPortraits~\cite{drobyshev2024emoportraits}}
& \multicolumn{4}{c}{LivePortrait~\cite{guo2024liveportrait}} \\
\cmidrule(lr){2-5}\cmidrule(lr){6-9}\cmidrule(lr){10-13}\cmidrule(lr){14-17}\cmidrule(lr){18-21}\cmidrule(lr){22-25}
Detector & AUC & T1 & T0.1 & Brier
         & AUC & T1 & T0.1 & Brier
         & AUC & T1 & T0.1 & Brier
         & AUC & T1 & T0.1 & Brier
         & AUC & T1 & T0.1 & Brier
         & AUC & T1 & T0.1 & Brier \\
\midrule
CADDM~\cite{dong2023implicit} & 0.82 & \cellcolor{colorpoor}{0.25} & \cellcolor{colorpoor}{0.21} & 0.023
& 0.98 & \cellcolor{colorpoor}{0.60} & \cellcolor{colorpoor}{0.60} & 0.06
& 0.95 & \cellcolor{colorpoor}{0.55} & \cellcolor{colorpoor}{0.20} & 0.11
& 0.93 & \cellcolor{colorpoor}{0.17} & \cellcolor{colorpoor}{0.17} & 0.10
& 0.75 & \cellcolor{colorpoor}{0.03} & \cellcolor{colorpoor}{0.03} & 0.24
& 0.95 & \cellcolor{colorpoor}{0.52} & \cellcolor{colorpoor}{0.41} & 0.09 \\
HiFi-Net~\cite{hifi_net_xiaoguo} & 0.88 & \cellcolor{colorpoor}{0.03} & \cellcolor{colorpoor}{0.01} & 0.17 & 0.86 & \cellcolor{colorpoor}{0.27} & \cellcolor{colorpoor}{0.11} & 0.16 & 0.84 & \cellcolor{colorpoor}{0.00} & \cellcolor{colorpoor}{0.00} & 0.18 & 0.48 & \cellcolor{colorpoor}{0.00} & \cellcolor{colorpoor}{0.00} & 0.44 & 0.61 & \cellcolor{colorpoor}{0.01} & \cellcolor{colorpoor}{0.00} & 0.34  & 0.84 & \cellcolor{colorpoor}{0.01} & \cellcolor{colorpoor}{0.00} & 0.27 \\
AltFreezing~\cite{wang2023altfreezing} & 1.00 & \cellcolor{colorreasonable}{0.87} & \cellcolor{colorfair}{0.78} & 0.03 & 1.00 & \cellcolor{colorreasonable}{0.94} & \cellcolor{colorreasonable}{0.88} & 0.03 & 1.00 & \cellcolor{colorgood}{0.95} & \cellcolor{colorreasonable}{0.89} & 0.02 & 0.99 & \cellcolor{colorreasonable}{0.85} & \cellcolor{colorpoor}{0.64} & 0.04 & 0.95 & \cellcolor{colorpoor}{0.58} & \cellcolor{colorpoor}{0.26} & 0.08 & 0.99 & \cellcolor{colorreasonable}{0.90} & \cellcolor{colorpoor}{0.56} & 0.03 \\
TALL~\cite{xu2023tall} & \textbf{1.00} & \cellcolor{colorexcellent}{\textbf{1.00}} & \cellcolor{colorexcellent}{\textbf{0.99}} & 0.01
& \textbf{1.00} & \cellcolor{colorexcellent}{\textbf{1.00}} & \cellcolor{colorexcellent}{\textbf{1.00}} & 0.01
& \textbf{1.00} & \cellcolor{colorexcellent}{0.99} & \cellcolor{colorexcellent}{\textbf{0.99}} & 0.01
& 0.97 & \cellcolor{colorfair}{0.84} & \cellcolor{colorfair}{\textbf{0.82}} & 0.06
& 0.97 & \cellcolor{colorpoor}{0.48} & \cellcolor{colorpoor}{0.48} & 0.15
& \textbf{1.00} & \cellcolor{colorreasonable}{0.93} & \cellcolor{colorreasonable}{\textbf{0.92}} & 0.02 \\
LipFD~\cite{liu2024lips} & 0.99 & \cellcolor{colorpoor}{0.72} & \cellcolor{colorpoor}{0.62} & 0.04
& 0.98 & \cellcolor{colorpoor}{0.71} & \cellcolor{colorpoor}{0.56} & 0.08
& 0.98 & \cellcolor{colorpoor}{0.72} & \cellcolor{colorpoor}{0.34} & 0.08
& 0.89 & \cellcolor{colorpoor}{0.57} & \cellcolor{colorpoor}{0.40} & 0.15
& 0.91 & \cellcolor{colorpoor}{0.30} & \cellcolor{colorpoor}{0.14} & 0.22
& 0.98 & \cellcolor{colorfair}{0.77} & \cellcolor{colorpoor}{0.61} & 0.06 \\
DF-Adapter~\cite{shao2025deepfake} & \textbf{1.00} & \cellcolor{colorreasonable}{0.93} & \cellcolor{colorpoor}{0.16} & 0.02
& \textbf{1.00} & \cellcolor{colorexcellent}{0.99} & \cellcolor{colorreasonable}{0.94} & 0.01
& \textbf{1.00} & \cellcolor{colorexcellent}{\textbf{1.00}} & \cellcolor{colorgood}{0.95} & 0.00
& \textbf{0.98} & \cellcolor{colorreasonable}{\textbf{0.88}} & \cellcolor{colorfair}{0.80} & 0.09
& \textbf{0.98} & \cellcolor{colorpoor}{\textbf{0.70}} & \cellcolor{colorpoor}{\textbf{0.53}} & 0.14
& \textbf{1.00} & \cellcolor{colorgood}{\textbf{0.95}} & \cellcolor{colorpoor}{0.56} & 0.03 \\
MM-Det~\cite{song2025learningmultimodalforgeryrepresentation} & 0.94 & \cellcolor{colorpoor}{0.49} & \cellcolor{colorpoor}{0.06} & 0.13 & 0.96 & \cellcolor{colorpoor}{0.40} & \cellcolor{colorpoor}{0.16} & 0.12 & 0.93 & \cellcolor{colorpoor}{0.33} & \cellcolor{colorpoor}{0.06} & 0.18 & 0.98 & \cellcolor{colorpoor}{0.55} & \cellcolor{colorpoor}{0.27} & 0.06 & 0.87 & \cellcolor{colorpoor}{0.06} & \cellcolor{colorpoor}{0.03} & 0.18 & 0.85 & \cellcolor{colorpoor}{0.08} & \cellcolor{colorpoor}{0.03} & 0.17 \\
\bottomrule
\end{tabular}}
\end{subtable}

\end{table*}

\noindent \textbf{Benchmarking Objectives and Key Questions} We perform a comprehensive evaluation and analysis to answer the following questions:\\
\noindent \textit{(i) How does the performance of the SOTA detectors change from FaceForensics++~\cite{rossler2019faceforensics++}, where detectors have near-perfect accuracy, to TalkingHeadBench?}\\
\noindent \textit{(ii) Can the SOTA detectors generalize across shifts in identity and generator property?}\\
\noindent \textit{(iii) Which talking-head generators are easy or hard to detect, and why?}\\
\noindent \textit{(iv) Which aspects of the TalkingHeadBench benchmark remain challenging for the SOTA detectors and warrant further investigation?}\\

\noindent \textbf{SOTA Detectors} To address these questions, we benchmark seven publicly available SOTA deepfake detectors: CADDM~\cite{dong2023implicit}, LipFD~\cite{liu2024lips}, DeepFake-Adapter~\cite{shao2025deepfake}, TALL~\cite{xu2023tall}, AltFreezing~\cite{wang2023altfreezing}, MM-Det~\cite{song2025learningmultimodalforgeryrepresentation}, and HiFi-Net~\cite{hifi_net_xiaoguo}.
Similar to the generators, we aim to diversify the modalities and frameworks of our detectors, as shown in Tab.~\ref{tab:detectors}. These seven detectors meet our goal of being the latest SOTA detectors with available code, and together, they capture varying modalities.

\noindent \textbf{Evaluation Protocols} To systematically evaluate the generalization of these SOTA detectors, we design three protocols:

\textit{P1 (Identity Shift Only) }Evaluates generalization to unseen identities within familiar generators. Models are trained on deepfakes from all generators and real videos, then tested on each generator and real videos, ensuring no overlap in identities between the training and test sets. (see Tab.~\ref{tab:m1_colored_updated}).

\textit{P2 (Generator Shift)} Evaluates generalization to an unseen generator. Models are trained on remaining generators and real videos, then tested on the held-out generator and real videos (see Tab.~\ref{tab:m2_scheme2_T1_colored}).

\textit{P3 (Identity \& Generator Shift)} Evaluates generalization to an unseen generator and identities. Using a leave-one-out approach, models are trained on the train splits of remaining generators and real videos, then tested on the test split of the held-out generator and real videos (see Tab.~\ref{tab:m3_scheme2_T1_colored}).

For these protocols, we select 1600 real videos from FaceForensics++~\cite{rossler2019faceforensics++} and CelebV-HQ~\cite{zhu2022celebv}, unseen during generation, to match fake video numbers per protocol, and perform face comparison to prevent identity leakage between train and test across real and fake video sets.

\noindent \textbf{Evaluation Metrics} Detector performance under these protocols is measured with three metrics: AUC, Brier Score, TPR@FPR=1\% (T1), and TPR@FPR=0.1\% (T0.1), following IJB-C face benchmark~\cite{maze2018iarpa}. Stricter thresholds on FPR, e.g., T0.1, are more useful when these detectors are applied at scale and only a small amount of bad detections are tolerable. We will prioritize T1 for most analysis in this paper, following its use in the IJB-C face benchmark, but later we will analyze the detector performance across various thresholds in FPR. We classify the performance of detectors as:  \\
Excellent: \colorbox{colorexcellent}{$\ge $0.99}, Good: \colorbox{colorgood}{0.95-0.99}, Reasonable: \colorbox{colorreasonable}{0.85-0.95}, Fair: \colorbox{colorfair}{0.75-0.85}, and Poor: \colorbox{colorpoor}{$<$0.75}.

\label{sec:analysis:tables}

\subsection{Impact of Generator and Identity Shifts}
\label{ssec:mode_analysis}

\begin{figure}[htbp]
    \centering
    \includegraphics[width=0.8\columnwidth]{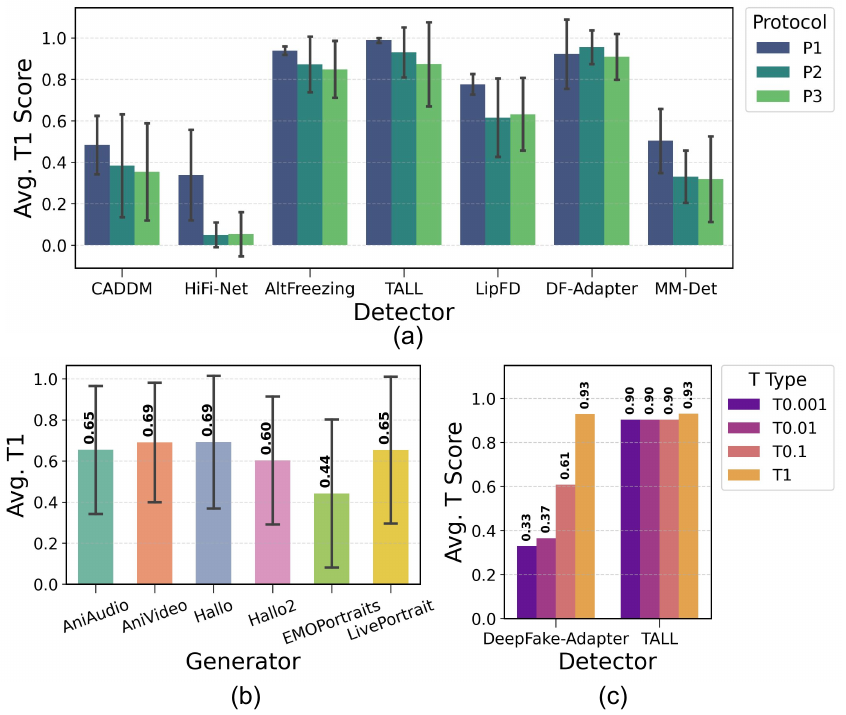}
    \captionsetup{font=small}
    \caption{Average performance scores across detectors and generators. 
    (a) shows the average T1 across all detectors for each evaluation protocol. Detector performance consistently declines from P1 to P2, and further in P3, highlighting the increasing difficulty of generalization. These results suggest that evaluating detectors solely on average performance may obscure generator-specific weaknesses. 
    (b) shows the average T1 score per generator across all detectors. EMOPortraits emerges as the most challenging generator, while the others fall within a narrower range of 0.60–0.69, indicating relatively comparable levels of detectability.
    (c) presents the average TPR at varying FPR thresholds for TALL and DeepFake-Adapter. TALL maintains high TPR even at extreme thresholds (e.g., FPR=0.001), demonstrating strong robustness. In contrast, DeepFake-Adapter exhibits a sharp performance drop as the threshold tightens, highlighting its reduced reliability under stricter operating conditions.}
    \label{fig:tpr_combined_all}
\end{figure}

\textbf{Protocol-Dependent Performance Shifts:} Overall, detectors exhibit performance shifts across protocols. In particular, average $AUC$ and T1 scores tend to decline from P1 to P2, indicating that generator shift (P2) poses greater challenges than identity shift (P1). This trend continues in P3, where combined generator and identity shifts further reduce performance as illustrated in Fig.~\hyperref[fig:tpr_combined_all]{\ref*{fig:tpr_combined_all}a}. For instance, TALL's~\cite{xu2023tall} average T1 score decreases from 0.99 in P1 to 0.93 in P2 and 0.87 in P3.
Despite these increasingly difficult conditions, TALL and DeepFake-Adapter~\cite{shao2025deepfake} maintain strong performance across all three protocols, with DeepFake-Adapter's lower score on P1 largely due to poor performance on AniPortrait Audio~\cite{wei2024aniportrait}(T1 = 0.58; see Tab.~\ref{tab:m1_colored_updated}). This underscores the need to assess detectors not just by averages, but also by generator-level sensitivity.

\textbf{Generator Difficulty and Cross-Protocol Generalization:}
Fig.~\hyperref[fig:tpr_combined_all]{\ref*{fig:tpr_combined_all}b} shows the variation of detector performance across generators.
EMOPortraits~\cite{drobyshev2024emoportraits} stands out as the most challenging case, with average T1 of only \textbf{0.44}, far below all others. By contrast, the remaining generators cluster within a narrower range of \textbf{0.60-0.69}, indicating more comparable levels of detectability. This sharp gap emphasizes that while detectors maintain moderate robustness across most generators, EMOPortraits remains a persistent failure case that reveals their limitations under stricter thresholds.

\subsection{Detector Robustness and SOTA Evaluation}
\label{ssec:robustness_SOTA}

\begin{figure}[htbp]
    \centering
    \includegraphics[width=\linewidth]{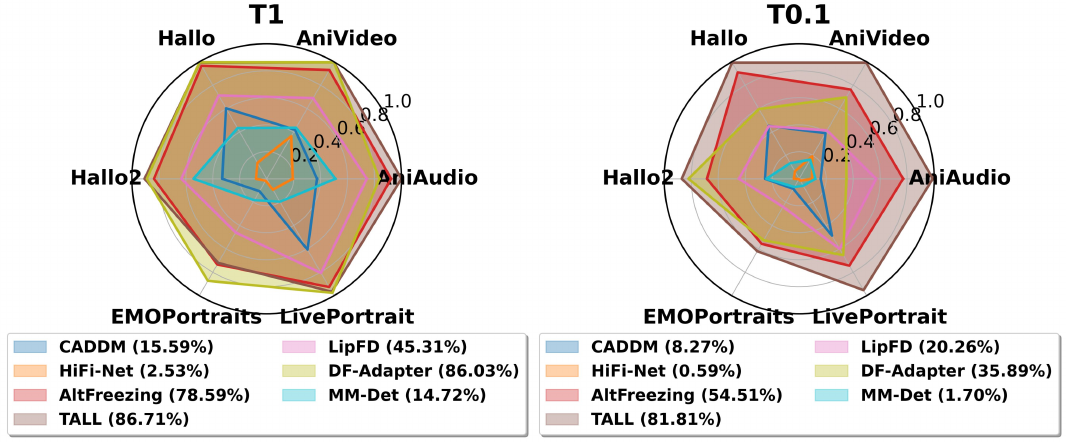}
    \captionsetup{font=small}
    \caption{Detector performance across generators measured by T1 and T0.1, averaged over all protocols. (a) shows T1 result, where DeepFake-Adapter and TALL demonstrate strong generalization across identity, generator, and joint shifts. (b) shows stricter T0.1 result, where most detectors have a noticeable drop, highlighting challenges in maintaining high recall under low FPR. Despite this, TALL remains the most robust, showing consistent performance across generators even under tighter operating conditions.}
    \label{fig:radar_plot_combined}
    \vspace{-3ex}
\end{figure}

\textbf{Overall Robustness Profile:} While detailed per protocol analysis is valuable, evaluating robustness across thresholds provides a comprehensive view of a detector’s generalization capabilities. Fig.~\ref{fig:radar_plot_combined} summarizes performance across all generators, averaged over protocols, for T1 and T0.1 thresholds. High and balanced coverage across all axes indicates consistent generalization across identity, generator, and joint shifts—crucial for handling unseen data from future SOTA generators and varying identities. At T1 (Fig.~\hyperref[fig:radar_plot_combined]{\ref*{fig:radar_plot_combined}a}), DeepFake-Adapter~\cite{shao2025deepfake} and TALL~\cite{xu2023tall} exhibit strong performance across most generators, with DeepFake-Adapter excelling on LivePortrait~\cite{guo2024liveportrait}, Hallo~\cite{xu2024hallo}, and AniPortraitVideo~\cite{wei2024aniportrait}. However, when the threshold is tightened to T0.1 (Fig.~\hyperref[fig:radar_plot_combined]{\ref*{fig:radar_plot_combined}b}), DeepFake-Adapter’s performance drops sharply—particularly for AniPortraitAudio~\cite{wei2024aniportrait} and EMOPortraits~\cite{drobyshev2024emoportraits}—indicating reduced robustness under stricter constraints. In contrast, TALL retains high scores across most generators at T0.1, suggesting better threshold-level resilience and stronger generalization under a low false positive rate. 
We also evaluate detector performance generalizing to unseen generators such as commercial generator MAGI-1~\cite{magi1} and academic generator Hallo3~\cite{cui2024hallo3}. We observe similar trend whereas the best-performing models, TALL and DeepFake-Adapter, maintain high detection scores. We include the full result in the Supplementary.

\textbf{Performance across different FPR thresholds:} While our primary evaluation focuses on T1 due to its importance in deepfake detection scenarios, analyzing stricter thresholds such as T0.1, T0.01, T0.001 also provides valuable insights into overall detector reliability. As shown in Fig.~\hyperref[fig:tpr_combined_all]{\ref*{fig:tpr_combined_all}c}, even though both TALL and DeepFake-Adapter achieve $\geq$ 0.90 for T1, while FPR decreases, we notice a significant drop for DeepFake-Adapter. This indicates its reduced reliability in real-world applications where minimizing such false accusations is paramount. Meanwhile, TALL demonstrates robust performance, maintaining a consistently high true positive rate even at these very low FPRs, highlighting its suitability for deployment in sensitive environments.  
 

\subsection{Generalization to Emerging Generators}
\label{ssec:genralize_gen}
To simulate a real-world scenario, we evaluated detectors against deepfakes from emerging academic generator Hallo3~\cite{cui2024hallo3} and commercial generator MAGI-1~\cite{magi1}, which were not seen during training. The results, provided in supplementary (Section 4), show that top-performing models like TALL and AltFreezing maintained high accuracy, suggesting that training on the diverse and curated set in TalkingHeadBench enables detectors to handle novel generator architectures. This indicates the detectors have learned a fundamental representation of talking-head artifacts beyond overfitting to specific generator fingerprints.

\subsection{Error Analysis and Explainability}
\label{ssec:error_analysis}
The quantitative results highlight areas requiring deeper investigation, for instance, the poor performance of most detectors against EMOPortraits~\cite{drobyshev2024emoportraits}, particularly in P3 where T1 scores drop significantly. Similarly, CADDM~\cite{dong2023implicit} underperforms consistently on TalkingHeadBench, indicating a need to examine its limitations more closely.

We leverage explainability techniques like Grad-CAM~\cite{Selvaraju_2019} to visualize the regions and features that detectors focus their predictions on. We demonstrate an example for TALL~\cite{xu2023tall} since it achieves the overall best performance on our dataset. Other detectors are in the supplementary.
 
\textbf{Failure Case Analysis:} On the EMOPortraits dataset in P3-TALL~\cite{xu2023tall}'s worst performing scenario (see Tab.~\ref{tab:m3_scheme2_T1_colored}), failure cases reveal that the model misclassifies both real and fake videos due to reliance on background cues. As shown in Fig.~\ref{fig:tall_failure_emop3}, real and fake videos are getting misclassified due to the model’s attention being disproportionately focused on the background, despite meaningful facial cues suggesting the correct label. This indicates a weakness in TALL's generalization: it is overly influenced by background artifacts, likely due to training bias, and fails to localize attention to the face under combined identity and generator shifts.


\begin{figure}[t]
  \centering
  \includegraphics[width=\linewidth]{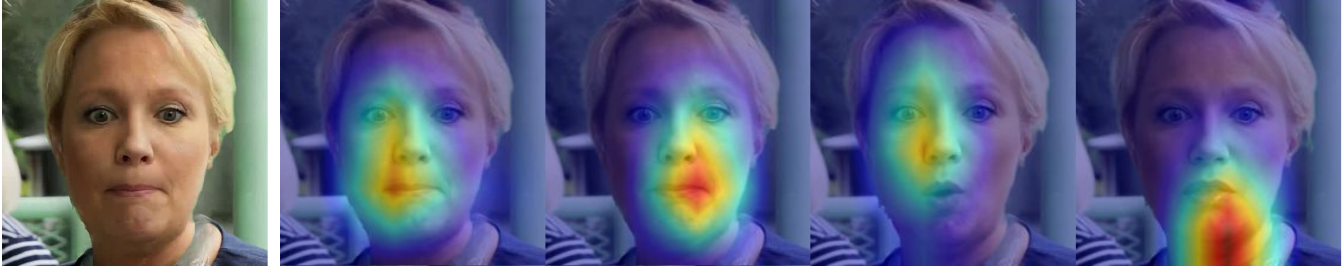}
  \caption{Success case of TALL on EMOPortrait (P3): the model correctly detects a deepfake by focusing on the neck, a known region of visual artifacts in EMOPortrait. This shows cross-dataset generalization despite the rarity of such artifacts in training data.}
  \label{fig:tall_success_emopm3}
\end{figure}

\textbf{Success Case Analysis:} In contrast to its failures, TALL demonstrates effective generalization in certain EMOPortraits~\cite{drobyshev2024emoportraits} videos under P3, as shown in Fig.~\ref{fig:tall_success_emopm3}. The model accurately classifies a deepfake by focusing on the neck region, which is known to exhibit artifacts in EMOPortraits. Notably, this success occurs despite such artifacts being uncommon in other datasets used during training. This suggests TALL~\cite{xu2023tall} has transferred knowledge from broader manipulation cues across datasets. The focused activation around the lower face and neck reflects meaningful model behavior and highlights its potential for cross-domain generalization.


\begin{table}[t]
  \centering
  \small
  \begin{tabular}{@{}lccc@{}}
    \toprule
    \textbf{Generator} & \textbf{TPs (P1)} & \textbf{FNs (P3)} & \textbf{\% Drop} \\
    \midrule
    EMOPortraits~\cite{drobyshev2024emoportraits} & 188 & 72 & 38.30\% \\
    Hallo2~\cite{cui2024hallo2}                   & 102 & 9  & 8.82\%  \\
    LivePortrait~\cite{guo2024liveportrait}       & 168 & 5  & 2.98\%  \\
    \bottomrule
  \end{tabular}
  \caption{Breakdown of high-confidence true positives (TPs) ($\geq 0.9$) in P1 that become false negatives (FNs) in P3 for TALL. A large drop from confident predictions in P1 to errors in P3 indicates poor generalization across generator shifts.}
  \label{tab:mode_shift_analysis}
\end{table}

\textbf{Tracking Generalization Failures from Confident Predictions:} To understand detector failures under distribution shifts, we track how high-confidence true positives (TPs) in P1 transition under P3. As shown in Tab.~\ref{tab:mode_shift_analysis}, for TALL—our best-performing detector—38.3\% of EMOPortraits samples confidently classified in P1 (Prediction $\geq$ 0.9) are misclassified as false negatives in P3. This drop suggests TALL struggles to generalize when both identity and generator properties change. In contrast, TALL shows only a 2.98\% drop from P1 to P3 on LivePortrait, highlighting that some generators pose far more severe generalization challenges. These results reinforce that P3 degradation is not random but rooted in generator-specific variation, with EMOPortraits emerging as a valuable stress test for detector evaluation.

\textbf{Protocol-Specific Feature Reliance:} 
To further understand detector behavior across generators, Fig.~\ref{fig:tall_heatmap_m1} presents Grad-CAM visualizations of TALL on samples from all six generators in Protocol 1, where we can study the bias of detectors to specific generator characteristics using the smallest train-test data distribution shift in P1. For generators such as AniPortrait (Audio/Video), Hallo, and LivePortrait, the model predominantly focuses on the central facial region, aligning well with human-interpretable deepfake cues. However, for Hallo2 and EMOPortraits, attention shifts away from facial features and toward areas surrounding the head. This may suggest: either the facial features in these generators are realistic enough to fool the detector, or the model has learned to rely on peripheral artifacts (e.g., spatial warping in Hallo2 or neck distortions in EMOPortraits) as key detection signals. Such behavior indicates both progress in generator realism and potential over-reliance on non-facial cues in certain detector architectures.
\begin{figure}[htbp]
  \vspace{-0.8em}
  \centering
  \includegraphics[width=\linewidth]{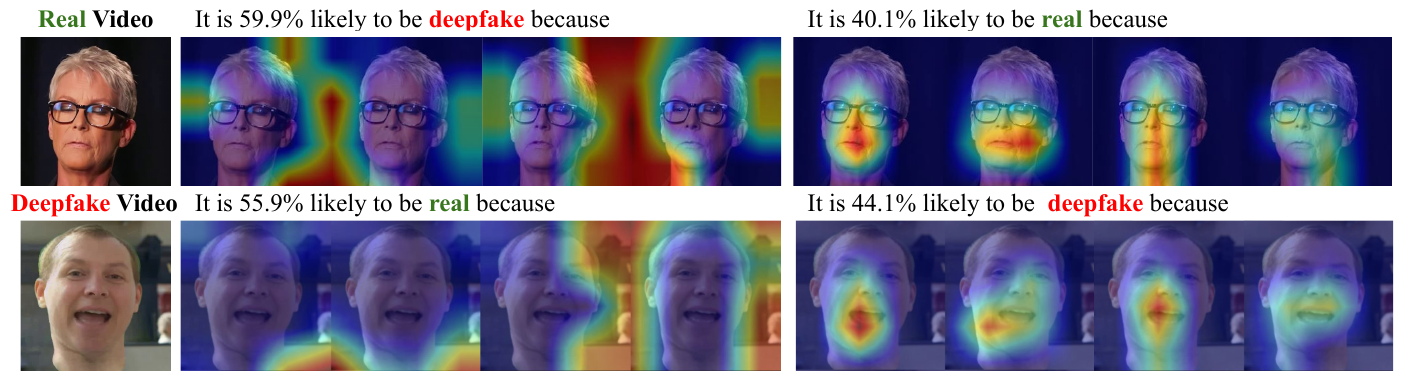}
  \captionsetup{font=small}
  \caption{Failure case from TALL on EMOPortraits (P3): misclassifications driven by attention to background regions rather than facial features. Despite correct facial cues, the model’s focus on irrelevant background areas leads to incorrect predictions.}
  \label{fig:tall_failure_emop3}
\end{figure}
\begin{figure}[htbp]
  \centering
  \includegraphics[width=\linewidth]{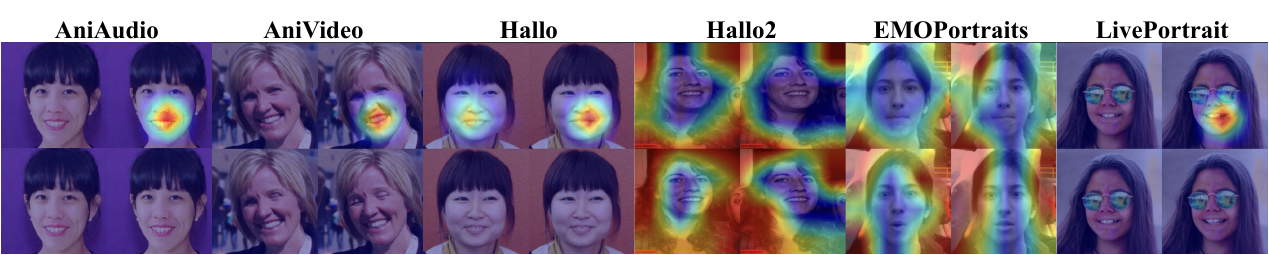}
  \captionsetup{font=small}
  \caption{Grad-CAM activation map from the TALL model on P1 samples across all six generators.}
  \label{fig:tall_heatmap_m1}
\end{figure}
\subsection{Guidance for Future Research using TalkingHeadBench}
\label{ssec:future_directions}
This benchmark underscores several critical areas for future research in deepfake detection.

\textbf{Limitations of SOTA Detectors.} Our analysis shows that while models like TALL~\cite{xu2023tall} and DeepFake-Adapter~\cite{shao2025deepfake} perform strongly across most generators, there remains room for improvement in achieving consistent excellence across domains. Specifically, TALL achieves exceptional T1 scores on 3 of 6 generators and no worse than fair on the others, while DeepFake-Adapter reaches exceptional on 2 and stays within reasonable to fair elsewhere. Notably, both models are jointly exceptional only on AniPortraitVideo~\cite{wei2024aniportrait} and Hallo~\cite{xu2024hallo}, suggesting detectors still struggle to generalize uniformly across diverse generation methods.

\textbf{Open Challenges.} While our analysis offers insights into robustness, several aspects of TalkingHeadBench remain difficult. These include (i) handling generators like EMOPortraits~\cite{drobyshev2024emoportraits} and Hallo2~\cite{cui2024hallo2}, where SOTA detectors reach only 70\% and 88\% T1 in P3; (ii) reducing degradation in P3 where both identity and generator differ between splits, as all detectors except DF-Adapter degrade significantly; (iii) overcoming reliance on background artifacts—as shown in Sec.~\ref{ssec:error_analysis}—which causes models to ignore relevant facial features. These challenges highlight the need for detectors with stronger domain adaptation, semantic attention, and modality-aware training to handle real-world deepfakes more reliably.

\textbf{Future Objective.} \\
(a) \textit{Performance:} The best model, TALL~\cite{xu2023tall}, reaches excellent T1 ($>99\%$) on only 3/6 generators in P3, while reaching $\sim$90\% for Hallo2 and LivePortrait and $\sim$80\% for EMOPortraits. Future research should strive for $>99\%$ in both T1 and T0.1 across all 6 detectors and protocols. \\
(b) \textit{Analysis:} TalkingHeadBench and its protocols offer the granularity to pinpoint weaknesses and guide improvements. Research should prioritize reducing drops in challenging generators (e.g., EMOPortraits, Hallo2), exploring domain adaptation, and enhancing sensitivity to subtle artifacts while avoiding overfitting to background cues. \\
(c) \textit{Evolving Benchmark:} A persistent challenge is that benchmarks lag behind generative models, quickly becoming outdated. While TalkingHeadBench on Hugging Face provides static splits, we aim to evolve it for both detector and generator communities. This includes (i) APIs for authors to test detectability, (ii) dual leaderboards to track progress, (iii) adding generators with lower detectability, and (iv) removing generators easily detectable ($>99\%$ T0.1) by most detectors every six months. This evolving benchmark will accelerate innovation and strengthen detection systems against newer generators.


By addressing these areas, the community can move towards deepfake detectors that are more accurate, robust, reliable, and trustworthy across diverse conditions encountered in real-world scenarios.

\section{Conclusions, Broad Impacts, and Limitations}
\label{sec:conclusions}



We introduce TalkingHeadBench, a comprehensive benchmark utilizing contemporary talking-head generators to stress-test deepfake detectors under realistic distribution shifts. Our evaluations reveal critical gaps: even state-of-the-art detectors struggle with generalization across changes in identity and generator, particularly with challenging fakes like EMOPortraits~\cite{drobyshev2024emoportraits}, often due to over-reliance on non-facial artifacts. TalkingHeadBench will broadly impact the community by providing a standardized, challenging benchmark that reflects modern deepfake realism and complexity. We will provide open access to the dataset, evaluation protocols, and baseline code.

While we acknowledge one potential risk for the benchmark is that it could facilitate more sophisticated talking-head generation techniques, which could be potentially misused for deceptive or malicious purposes, our primary objective is to catalyze research into more robust, generalizable, and artifact-aware detection methods. 

The dataset only contains a finite number of videos and identities derived from FFHQ~\cite{karras2019style} and CelebV-HQ~\cite{zhu2022celebv}, which might not represent the diversity of real-world deepfakes that can be exhibited.
While we focus on creating high-quality forgeries, we acknowledge that our dataset can inherit the demographic imbalances from the public source datasets. We provide a detailed demographic audit in the Supplementary for transparency, and identify notable
disparities across racial and age groups, while gender bias is minimal.
In the future, we will set up leaderboards that allow performance comparison among newly developed generators and detectors.
Additionally, we will conduct regular fairness audits, applying both group-based and individual fairness metrics to ensure robustness in our future leaderboard as new generators emerge.
This benchmark serves as a crucial resource for developing trustworthy deepfake detection systems, essential for mitigating the increasing societal risks posed by manipulated media.

\clearpage
\section{Acknowledgement}
\label{sec:acknowledgement}

This research was supported in part by Lenovo Research (Morrisville, NC). We gratefully acknowledge the invaluable support and assistance of the members of the Mobile Technology Innovations Lab. 
{
    \small
    \bibliographystyle{ieeenat_fullname}
    \bibliography{references, references_luchao}
}

\clearpage
\appendix
\section{Supplementary Materials}


All benchmarks and curated dataset for our TalkingHeadBench are publicly released at \href{https://anaxqx.github.io/talkingheadbench.github.io}{https://anaxqx.github.io/talkingheadbench.github.io}.
TalkingHeadBench is released under a Creative Commons Attribution 4.0 License (CC BY 4.0).

Our supplementary materials are summarized as follows:
\begin{itemize}
    \item \textbf{Details of Assets Used (Section \ref{suppl-sec:asset_details}):} Descriptions, sources, and licenses for source datasets, talking-head generators, and deepfake detectors utilized in our benchmark.
    \item \textbf{Detailed Experimental Setup (Section \ref{suppl-sec:detailed_experimental_setup}):} Dataset splits, data curation procedures, computational resources, and key training hyperparameters.
    \item \textbf{Detailed Per-Generator Artifact Observations (Section \ref{suppl-sec:artifact_observations}):} Specific types of visual artifacts commonly produced by each generator, noted during our manual curation process.
    \item \textbf{Experimental Results (Section \ref{suppl-sec:additional_results}):} Performance for commercial generator (MAGI-1) and academic generator (Hallo3).
    \item \textbf{Explainability Results (Section \ref{suppl-sec:additional_gradcam}):} Grad-CAM visualizations for benchmarked detectors.
\end{itemize}

\section{Details of Assets Used}
\label{suppl-sec:asset_details}

\subsection{Source Datasets for Generation and Real Videos}
\label{suppl-subsec:source_datasets}
TalkingHeadBench leverages publicly available datasets for source images and driving signals (audio/video) for talking-head generation and real videos for detector training and testing.

\paragraph{Source Images for Deepfake Generation}
\begin{itemize}
    \item \textbf{FFHQ (Flickr-Faces-HQ)~\cite{karras_style-based_2019}:} This dataset served as the primary source for high-quality, diverse static portrait images used as the base for deepfake generation. It contains 70,000 high-resolution images of diverse human faces with diverse ages, ethnicities, and image backgrounds originally intended for GAN benchmarking. We utilized images from the 1024x1024 resolution subset. The dataset can be accessed through the following link: \url{https://github.com/NVlabs/ffhq-dataset} under Creative Commons BY-NC-SA 4.0 license by NVIDIA Corporation.
\end{itemize}

\paragraph{Source Driving Signals (Audio/Video) for Deepfake Generation}
\begin{itemize}
    \item \textbf{CelebV-HQ~\cite{zhu2022celebv}:} This large-scale video facial attributes dataset was used as the source for dynamic driving signals. It contains a variety of identities, ethnicities, expressions, and poses. Both video segments (for video-driven generators) and extracted audio tracks (for audio-driven generators, converted to .wav format) were utilized. The dataset can be accessed through the following link: \url{https://github.com/CelebV-HQ/CelebV-HQ}. The copyright information for this dataset is not explicitly mentioned. 
\end{itemize}

\paragraph{Real Videos for Detector Training/Testing}
\begin{itemize}
    \item \textbf{FaceForensics++ (FF++)~\cite{rossler2019faceforensics++}:} Original, unmanipulated video sequences were included in our set of real videos.  We directly downloaded the dataset following the download script they provided, and finalized the number of our YouTube real videos from 1000 to 704 due to the unavailability of some of the videos on YouTube. We then renamed the downloaded audio to match the naming of the YouTube real videos. The renamed audio files are in our Hugging Face datasets (./audio/ff++).  The original FF++ dataset can be accessed through the following link: \url{https://github.com/ondyari/FaceForensics}. The copyright information can be found at \url{https://github.com/ondyari/FaceForensics/blob/master/LICENSE}. 
    \item \textbf{CelebV-HQ~\cite{zhu2022celebv}:} To diversify the identity in our real videos, we incorporated a distinct set of real videos from CelebV-HQ that were not used for driving signal extraction.
    \item \textbf{Identity Control:} To prevent identity leakage and ensure fair evaluation, rigorous identity separation was maintained. Identities in the real video sets did not overlap with those used for generating deepfakes (either source images or driving signals) or across the train/test splits of the real videos themselves. This was verified using face recognition (InsightFace with ArcFace model on middle frames of videos).
\end{itemize}

\subsection{Talking-Head Generators}
\label{suppl-subsec:generators_details}
The deepfake videos in TalkingHeadBench were generated using the following state-of-the-art talking-head generators. We provide brief descriptions, primary modality, and links to code repositories/licenses where available. Visual examples are in Fig. 1.


\paragraph{Hallo~\texorpdfstring{\cite{xu2024hallo}}{[xu2024hallo]}}
\begin{itemize}
    \item \textbf{Description:} Hallo is a talking-head generator that was published in 2024. The generator aims to create realistic and temporally consistent talking-head animations from a static portrait image and a driving audio clip. Hallo integrates diffusion-based models, a UNet-based denoiser, temporal alignment techniques, and a reference network. This hierarchical audio-driven method allows for diversity in poses and expressions. This end-to-end approach enhances animation quality, motion diversity, and personalization across different identities.
    \item \textbf{Modality:} Audio-driven.
    \item \textbf{Code Repository:} \url{https://github.com/fudan-generative-vision/hallo} 
    \item \textbf{License:} MIT License. 
    \item \textbf{Key Artifacts Observed:} See Section \ref{subsubsec:hallo_artifacts}.
\end{itemize}

\paragraph{Hallo2~\texorpdfstring{\cite{cui2024hallo2}}{[cui2024hallo2]}}
\begin{itemize}
    \item \textbf{Description:} Hallo2 is an audio-driven talking-head generator that extends Hallo by enabling long-duration and high-resolution video synthesis published in ICLR2025. It incorporates enhanced temporal modeling and spatial fidelity mechanisms to produce stable and expressive animations over extended sequences. Built upon a diffusion-based generation backbone, Hallo2 introduces improved temporal alignment and rendering strategies that support better lip-sync accuracy and identity preservation. Hallo2 is suited for real-world applications such as virtual avatars, video dubbing, and personalized content creation.
    \item \textbf{Modality:} Audio-driven.
    \item \textbf{Code Repository:} \url{https://github.com/fudan-generative-vision/hallo2} 
    \item \textbf{License:} MIT License. 
    \item \textbf{Key Artifacts Observed:} See Section \ref{subsubsec:hallo2_artifacts}.
\end{itemize}

\paragraph{Hallo3~\texorpdfstring{\cite{cui2024hallo3}}{[cui2024hallo3]}}
\begin{itemize}
    \item \textbf{Description:} Hallo3 is an audio-driven portrait image animation model that uses a diffusion transformer backbone, with an identity reference network combining a causal 3D VAE plus stacked transformer layers, to generate highly dynamic, realistic video from static portraits. It improves over U-Net based generators by better handling non-frontal perspectives, immersive backgrounds, and motion dynamics, while preserving identity consistency even across viewpoint changes.
    \item \textbf{Modality:} Audio-driven.
    \item \textbf{Code Repository:} \url{https://github.com/fudan-generative-vision/hallo3} 
    \item \textbf{License:} MIT License. 
    \item \textbf{Key Artifacts Observed:} Generally high quality; specific subtle artifacts, if any, are less pronounced or systematic compared to some open-source academic models.
\end{itemize}

\paragraph{AniPortrait~\texorpdfstring{\cite{wei2024aniportrait}}{[wei2024aniportrait]}}
\begin{itemize}
    \item \textbf{Description:} 
    AniPortrait is a talking-head generator introduced in 2024. The two-stage approach first extracts 3D representations from audio, converting them to 2D facial landmark sequences. Then, a diffusion model with a motion module transforms these landmarks into temporally coherent talking-head videos. The hierarchical audio-driven pipeline offers precise control over audio-driven facial expressions and head poses while maintaining identity through reference guidance. Its training scheme, which decouples identity encoding from motion dynamics, enables an end-to-end approach that produces animations with accurate lip-sync, detailed expressions, and robust identity preservation. In video-driven mode, AniPortrait directly extracts facial landmarks from a source video to transfer expressions and movements to the reference portrait, utilizing the same reenactment pipeline for consistent temporal guidance and high-quality results.
    \item \textbf{Modality:} Audio-driven or video-driven.
    \item \textbf{Code Repository:} \url{https://github.com/Zejun-Yang/AniPortrait} 
    \item \textbf{License:} Apache-2.0 license. 
    \item \textbf{Key Artifacts Observed:} See Section \ref{subsubsec:aniportrait_audio_artifacts} for audio-driven artifacts and \ref{subsubsec:aniportrait_video_artifacts} for video-driven artifacts.
\end{itemize}

\paragraph{LivePortrait~\texorpdfstring{\cite{guo2024liveportrait}}{[guo2024liveportrait]}}
\begin{itemize}
    \item \textbf{Description:} LivePortrait is a real-time talking-head generator designed to produce high-quality portrait animations from a single static image introduced in 2024. It employs an implicit keypoint-based architecture combined with lightweight retargeting and stitching modules to ensure accurate facial motion and head pose transfer. Trained on a large-scale dataset of over 69 million frames, LivePortrait generalizes well across diverse identities and expressions.
    \item \textbf{Modality:} Video-driven.
    \item \textbf{Code Repository:} \url{https://github.com/KwaiVGI/LivePortrait} 
    \item \textbf{License:} MIT License. 
    \item \textbf{Key Artifacts Observed:} See Section \ref{subsubsec:liveportrait_artifacts}.
\end{itemize}

\paragraph{EMOPortraits~\texorpdfstring{\cite{drobyshev2024emoportraits}}{[drobyshev2024emoportraits]}}
\begin{itemize}
    \item \textbf{Description:} EMOPortraits is a one-shot talking-head generator introduced in CVPR2024. It employs a two-stage training process, with an optional audio-driven phase for video generation from a single image and audio input. The model selects two random frames of the source and driver at each step, adapts the driver frame's motion and expressions onto the source frame to generate the final image. It encodes a source image into a 3D latent feature and identity descriptor, while a motion module extracts pose/expression codes from a driver, resulting in realistic deepfakes under extreme and asymmetric facial expressions.
    \item \textbf{Modality:} Video-driven.
    \item \textbf{Code Repository:} \url{https://github.com/neeek2303/EMOPortraits} 
    \item \textbf{License:} Apache-2.0 license. 
    \item \textbf{Key Artifacts Observed:} See Section \ref{subsubsec:emoportraits_artifacts}.
\end{itemize}

\paragraph{MAGI-1~\texorpdfstring{\cite{magi1}}{[magi1]} (Commercial)} 
\begin{itemize}
    \item \textbf{Description:} MAGI-1 is a talking-head generator published in 2025. This open-source generator produces realistic, high-quality, temporally consistent videos from text or image prompts. Built on a diffusion transformer architecture, MAGI-1 generates fixed-length videos, enabling real-time streaming and seamless video continuation. With support for large-scale model sizes and long context lengths, it is well-suited for a wide range of creative and generative video applications.
    \item \textbf{Modality:} Primarily Text-to-Video, can optionally add a reference image.
    \item \textbf{Code Repository:} \url{https://github.com/SandAI-org/MAGI-1}
    \item \textbf{License:} Apache-2.0 license. 
    \item \textbf{Key Artifacts Observed:} Generally high quality; specific subtle artifacts, if any, are less pronounced or systematic compared to some open-source academic models. Focus is often on overall scene coherence rather than just facial animation.
\end{itemize}

\subsection{DeepFake Detectors}
\label{suppl-subsec:detectors_details}
The following SOTA deepfake detection models were benchmarked on TalkingHeadBench. Brief descriptions, primary input modality, and links to code repositories/licenses are provided.


\paragraph{CADDM~\texorpdfstring{\cite{dong2023implicit}}{[dong2023implicit]}}
\begin{itemize}
    \item \textbf{Description:} CADDM utilizes an Artifact Detection Module designed to focus on local regions of images. This module employs multiscale anchors to detect and classify artifact areas, aiming to mitigate the influence of identity information in deepfake detection, making it generalizable across different datasets. It showed great improvement in both accuracy and robustness when evaluating FF++. We directly deployed the model from the official GitHub repository to our cluster without modifying any hyperparameter or model structure.
    \item \textbf{Modality:} Image-based (CNN).
    \item \textbf{Code Repository:} \url{https://github.com/megvii-research/CADDM} 
    \item \textbf{License:} Apache-2.0 license. 
\end{itemize}

\paragraph{TALL~\texorpdfstring{\cite{xu2023tall}}{[xu2023tall]}}
\begin{itemize}
    \item \textbf{Description:} TALL introduces a temporal-attentive localization and learning framework designed to exploit temporal inconsistencies in deepfake videos. By leveraging a dual-stream architecture that models both short-term and long-term temporal dependencies, TALL isolates manipulated frames and emphasizes temporal transitions typically ignored by frame-based detectors. A temporal attention mechanism further enhances frame-level representations by focusing on abrupt motion irregularities introduced by forgery generation processes. Moreover, the authors propose TALL-Swin, which integrates the TALL strategy with the Swin Transformer architecture. This combination leverages the Swin Transformer's hierarchical feature representation and shifted window mechanism to effectively model both local and global dependencies within the thumbnail layout. 
    Given there is no available model released on their original GitHub repository, we pretrained the model using FF++ based on the implementation version in DeepfakeBench~\cite{yan_deepfakebench_2023}.
    \item \textbf{Modality:} Video-based (Transformer).
    \item \textbf{Code Repository:} \url{https://github.com/rainy-xu/TALL4Deepfake} 
    \item \textbf{License:} MIT license.
\end{itemize}

\paragraph{LipFD~\texorpdfstring{\cite{liu2024lips}}{[liu2024lips]}}
\begin{itemize}
    \item \textbf{Description:} LipFD targets lip-sync inconsistency by focusing on the alignment between lip motion and audio speech content. The model extracts fine-grained spatio-temporal features of mouth regions and correlates them with phoneme-level audio embeddings to detect subtle mismatches indicative of forgery. By isolating this cross-modal inconsistency, LipFD distinguishes itself from generic detectors that rely primarily on visual features. This focused approach leads to robust detection of audio-driven deepfakes, especially under real-world conditions. We directly reproduced the results from official GitHub repository with minimal adjustments on learning rate and learning rate decay. 
    \item \textbf{Modality:} Audio-Visual (Transformer).
    \item \textbf{Code Repository:} \url{https://github.com/AaronComo/LipFD} 
    \item \textbf{License:} N/A 
\end{itemize}

\paragraph{DeepFake-Adapter~\texorpdfstring{\cite{shao2025deepfake}}{[shao2025deepfake]}}
\begin{itemize}
    \item \textbf{Description:} DeepFake-Adapter presents a universal adaptation framework for deepfake detection by leveraging modality specific adapters integrated into a unified transformer backbone. These adapters specialize in capturing subtle, forgery-specific artifacts across diverse data modalities (e.g., RGB, Depth, Frequency), enabling cross-modal generalization without retraining. We directly reproduced the results from official GitHub repository with minimum adjustments to train using our custom dataset.
    \item \textbf{Modality:} Image-based (Transformer).
    \item \textbf{Code Repository:} \url{https://github.com/rshaojimmy/DeepFake-Adapter}
    \item \textbf{License:} N/A 
\end{itemize}

\paragraph{AltFreezing~\texorpdfstring{\cite{wang2023altfreezing}}{[wang2023altfreezing]}}
\begin{itemize}
    \item \textbf{Description:} A video-based deepfake detector that leverages a spatiotemporal model. It employs a training strategy for 3D ConvNet video detectors that alternately freezes spatial- and temporal-related parameter groups to force learning of both artifact types, yielding stronger out-of-distribution generalization on face forgery videos.
    \item \textbf{Modality:} Video-based (3D ConvNet / spatiotemporal).
    \item \textbf{Code Repository:} \url{https://github.com/ZhendongWang6/AltFreezing}
    \item \textbf{License:} MIT license. 
\end{itemize}

\paragraph{MM-Det~\texorpdfstring{\cite{song2025learningmultimodalforgeryrepresentation}}{[song2025learningmultimodalforgeryrepresentation]}}
\begin{itemize}
    \item \textbf{Description:} A diffusion deepfake video detector that builds a Multi-Modal Forgery Representation (MMFR) using a large multi-modal model (LLaVA) and fuses it with a spatiotemporal backbone enhanced by In-and-Across Frame Attention (IAFA). Introduces the DVF dataset and reports SOTA on diffusion-generated videos.
    \item \textbf{Modality:} Video-based (Diffusion).
    \item \textbf{Code Repository:} \url{https://github.com/SparkleXFantasy/MM-Det}
    \item \textbf{License:} Apache-2.0 license. 
\end{itemize}

\paragraph{HiFi-Net~\texorpdfstring{\cite{hifi_net_xiaoguo}}{[hifi_net_xiaoguo]}}
\begin{itemize}
    \item \textbf{Description:} An image forgery detection \& localization framework with a hierarchical fine-grained formulation: multi-branch feature extractor plus dedicated localization and classification heads to capture subtle artifacts across manipulation types.
    \item \textbf{Modality:} Image-based (detection + localization).
    \item \textbf{Code Repository:} \url{https://github.com/CHELSEA234/HiFi_IFDL}
    \item \textbf{License:} MIT license. 
\end{itemize}

\section{Detailed Experimental Setup}
\label{suppl-sec:detailed_experimental_setup}

\subsection{Dataset Generation, Splits, and Curation Details}
\label{suppl-subsec:data_splits_curation}

\paragraph{Deepfake Video Generation}
Approximately 500-600 videos were initially generated for each of the six open-source academic models. This was achieved by scripting random pairings of source images from FFHQ~\cite{karras_style-based_2019} with driving signals (audio or video) from distinct splits of CelebV-HQ~\cite{zhu2022celebv}.

\paragraph{Data Curation and Quality Control}
A rigorous manual curation process was undertaken to ensure the quality and challenge of the benchmark. This involved:
\begin{itemize}
    \item Reviewing each generated video for visual fidelity and realism.
    \item Removing videos with obvious generation failures, extreme distortions, or artifacts that make them trivially identifiable as fake (unless such artifacts are characteristic and subtle).
    \item Ensuring that the remaining fakes posed a reasonable challenge to detection models.
\end{itemize}
This process resulted in the final dataset sizes reported in Tab.4 of the main paper, with approximately 60-65\% of initially generated videos being discarded.

\paragraph{Train/Test Splits and Real Data Integration}
\begin{itemize}
    \item \textbf{Identity Separation:} Strict identity separation was enforced for all data. Source identities from FFHQ and driving signal identities from CelebV-HQ used for the training set of deepfakes were disjoint from those used for the test set.
    \item \textbf{Real Videos:} Real videos from FF++ and CelebV-HQ were also split into training and testing sets, maintaining identity separation. The number of real videos was balanced to be approximately 1:1 with fake videos in the training splits for each protocol.
    \item \textbf{Validation Sets:} For model validation during training, a small subset of identities from the training pool (both real and fake) was held out. For Protocol 1, this involved 50 real and 50 fake videos. For Protocols 2 and 3, it was 50 real and 50 fake videos per generator.
\end{itemize}

\subsection{Computational Resources}
\label{suppl-subsec:computational_resources}

The generation of TalkingHeadBench and the benchmarking of detection models were conducted using NVIDIA RTX A4500 GPUs. Deepfake generation times varied significantly depending on the generator model and the number of GPUs used, as videos were generated in parallel across multiple GPUs. On average, producing 500 videos took several hours to up to a day. For detector training, most state-of-the-art models completed training within 4 hours on a single GPU, with the exception of one model that required approximately an hour per epoch—making its total training time dependent on the specific protocol used.


\subsection{Hyperparameters and Training Details for Detectors}
\label{suppl-subsec:hyperparameters}
For all benchmarked detectors, we adhered closely to the training configurations and hyperparameters proposed in their original publications and official codebases. 

\section{Detailed Per-Generator Artifact Observations}
\label{suppl-sec:artifact_observations}
This section details characteristic visual artifacts observed for each generator during the manual data curation phase. These insights informed our curation and highlight the unique challenges posed by different generation techniques. 

\subsection{LivePortrait~\texorpdfstring{\cite{guo2024liveportrait}}{[guo2024liveportrait]}}
\label{subsubsec:liveportrait_artifacts}
Common artifacts observed in generations from the LivePortrait model included:
\begin{itemize}
\item \textbf{Non-Physical Head Kinematics and Scaling:} Driving videos featuring substantial translational or lateral head motion often induced unnatural scaling (e.g., apparent growth or shrinkage) or other non-physical movements in the synthesized head, indicative of inconsistent head movement/pose and face warping/distortion.
\item \textbf{Sensitivity to Source Image Composition:} The generation process exhibited heightened susceptibility to artifacts when source images contained non-adult subjects or multiple individuals. In scenarios with multiple people, artifacts manifested as face warping/distortion, erroneous animation of partially occluded or out-of-frame figures, and visible errors related to internal bounding box estimations, leading to significant spatial inconsistencies.
\item \textbf{Semantic Misinterpretation and Occlusion Errors:} Objects proximal to the head in the source image were sometimes erroneously segmented as facial features, resulting in their co-movement with the head, a form of occlusion error or texture anomaly.
\item \textbf{Lip Synchronization Deficiencies with Extreme Expressions:} Source images depicting pronounced oral expressions (e.g., open mouths, broad smiles, compressed lips) occasionally resulted in static labial regions or aberrant lip articulation, indicating lip sync errors or failures in modeling extreme facial deformations.
\end{itemize}

\subsection{Hallo~\texorpdfstring{\cite{xu2024hallo}}{[xu2024hallo]}}
\label{subsubsec:hallo_artifacts}
Common artifacts observed in generations from the Hallo model included:
\begin{itemize}
\item \textbf{Lip Synchronization Discrepancies:} Generated videos frequently exhibited a temporal mismatch between the audible speech and the synthesized lip movements, a common form of lip sync error.
\item \textbf{Hair Rendering Artifacts:} Portions of the synthesized hair often displayed unnatural stasis, appeared to adhere to the background, or were rendered with low fidelity (e.g., hair artifacts), particularly noticeable during dynamic head movements. This can be considered a type of temporal inconsistency or texture anomaly.
\item \textbf{Occlusion Handling Deficiencies:} The model demonstrated improper rendering when facial regions were expected to be occluded by elements such as eyeglasses, hair, or hands (if present in the source), leading to occlusion errors.
\item \textbf{Background Distortion:} The background region immediately surrounding the synthesized head was prone to background warping/distortion correlating with the head movement.
\end{itemize}

\subsection{Hallo2~\texorpdfstring{\cite{cui2024hallo2}}{[cui2024hallo2]}}
\label{subsubsec:hallo2_artifacts}
Generations from Hallo2 exhibited artifacts similar to its predecessor, Hallo, albeit with some distinct manifestations:
\begin{itemize}
\item \textbf{Aberrant or Absent Lip Articulation:} Instances of absent lip movement despite audible speech, or significant asynchrony between audio and visual lip cues, were noted, representing pronounced lip sync errors and temporal inconsistencies.
\item \textbf{Static Peripheral Hair Artifacts:} Hair situated outside the primary head bounding box frequently remained static during head motion, creating a visually jarring "detached" effect—a specific type of hair artifact and temporal inconsistency.
\item \textbf{Object-Induced Spatial Distortion:} The presence of objects proximate to or intersecting with the head's bounding box in the source image often precipitated localized face warping/distortion or spatial inconsistencies in the output.
\item \textbf{Background-Correlated Visual Anomalies:} Unsystematic visual artifacts, such as unpredictable patterns or noise, occasionally manifested, appearing to be correlated with complex textures or patterns within the source image's background, a form of spatial inconsistency or texture anomaly.
\end{itemize}

\subsection{AniPortrait~\texorpdfstring{\cite{wei2024aniportrait}}{[wei2024aniportrait]}}
\label{subsubsec:aniportrait_artifacts}
\paragraph{Audio-driven}
\label{subsubsec:aniportrait_audio_artifacts}
Analysis of audio-driven outputs from AniPortrait revealed the following:
\begin{itemize}
\item \textbf{Variable Perceptual Realism:} The model demonstrated capacity for producing naturalistic results, particularly notable for an exclusively audio-driven synthesis approach.
\item \textbf{Synchronization and Kinematic Irregularities:} The system was prone to occasional lip sync errors and exhibited unnatural head kinematics, including sudden accelerations, oscillatory movements, or an overly rigid, static posture, all categorized under inconsistent head movement/pose.
\item \textbf{Dental Structure Anomalies:} Synthesized dentition sometimes presented with unrealistic characteristics, such as supernumerary or atypically arranged rows of teeth (teeth anomalies).
\item \textbf{Pupillary Artifacts with Eyewear:} Source images featuring subjects with eyeglasses occasionally led to anomalous pupillary dilation patterns, such as variable frequencies of dilation (pupil anomalies).
\item \textbf{Hand-Induced Artifacts:} The presence of hands within the source image frame was a primary trigger for various occlusion errors and spatial inconsistencies.
\end{itemize}

\paragraph{Video-driven}
\label{subsubsec:aniportrait_video_artifacts}
Video-driven synthesis using AniPortrait was characterized by:
\begin{itemize}
\item \textbf{Microphone Occlusion Handling:} The model displayed an unusual proficiency in generating plausible outputs when the driving video featured a subject with a microphone in close proximity to the mouth, suggesting effective handling of this specific occlusion scenario.
\item \textbf{Catastrophic "Head Explosion" Artifacts:} A frequent failure mode involved severe face warping/distortion, where the synthesized head appeared to disintegrate or become overlaid with disparate visual elements (often misidentified hands or other body parts) from the driving video, a critical spatial inconsistency.
\item \textbf{Pervasive Background Instability:} Background warping/distortion was commonly found even in outputs that were otherwise subjectively assessed as high quality.
\item \textbf{Motion Tracking Fidelity versus Distortion Thresholds:} While the model effectively tracked subtle head movements, more pronounced motions (e.g., swaying) frequently surpassed its stable generation threshold, resulting in severe face warping/distortion that could occupy a significant portion of the video frame, indicating limitations in maintaining inconsistent head movement/pose coherence.
\item \textbf{Hair Segmentation Artifacts:} In certain instances involving female source images, synthesized hair was subject to abrupt truncation or unnatural rendering.
\item \textbf{Severe Hand-Related Artifacts:} The model exhibited a notable inability to process hand movements in the driving video, almost invariably leading to prominent and clearly delineated occlusion errors and spatial inconsistencies.
\item \textbf{High Incidence of Severe Generation Failures:} A substantial proportion of initial generations were unusable due to extreme face warping/distortion, transforming the source image into an unrecognizable amalgam of textures and features, indicative of fundamental failures in identity preservation and coherent image synthesis.
\end{itemize}

\subsection{EMOPortraits~\texorpdfstring{\cite{drobyshev2024emoportraits}}{[drobyshev2024emoportraits]}}
\label{subsubsec:emoportraits_artifacts}
Artifacts observed in EMOPortraits generations included:
\begin{itemize}
\item \textbf{"Floating Head" Artifacts:} Facial and head modifications were often restricted to the cephalic region, leading to a dissociation from neck and body movements. Significant corporeal motion in the driving video could result in the head appearing detached or "floating," a clear example of inconsistent head movement/pose and spatial inconsistency.
\item \textbf{Perifacial Blurring and Edge Anomalies:} The model frequently introduced blurring in the regions immediately surrounding the synthesized face, potentially diminishing perceptual realism and constituting an edge anomaly or texture anomaly.
\item \textbf{Off-Axis Pose Instability:} The system encountered difficulties rendering non-frontal (e.g., three-quarter or full profile) views. Lateral head rotations could precipitate severe face warping/distortion, including apparent shrinkage, color mismatch or degradation into noise resembling video static.
\item \textbf{Eyewear-Related Artifacts:} The model struggled with subjects wearing eyeglasses, particularly sunglasses. This often resulted in occlusion errors or texture anomalies where eyes were unrealistically rendered as visible through otherwise opaque lenses, or pupil anomalies.
\item \textbf{Chroma Key-like Color Artifacts:} Generated videos frequently exhibited spurious patches of bright, uniform color (commonly green), reminiscent of poorly executed chroma keying, indicating significant color mismatch or spatial inconsistencies.
\end{itemize}

\section{Additional Experimental Results}
\label{suppl-sec:additional_results}

\subsection{MAGI-1 Commercial Generator Results}
\label{suppl-subsec:magi1_results}

We present detection performance on deepfakes generated by the MAGI-1 commercial model, an unseen generator included to test detector robustness against proprietary systems. Since MAGI-1 does not provide audio output, we restrict evaluation to detectors operating on visual input: CADDM, HiFi-Net, TALL, and DeepFake-Adapter (DF-Adapter). Results are reported in Tab.~\ref{tab:mag1_t1}.

Overall, \textbf{TALL} achieves perfect scores across all metrics (AUC, T1, T0.1), indicating strong robustness to MAGI-1’s generation pipeline. \textbf{DF-Adapter} also performs near-perfectly, though its performance drops slightly at stricter thresholds (e.g., T0.1 falling to 0.99 on some splits). By contrast, \textbf{CADDM} achieves only moderate performance (AUC 0.88–0.97, T1 around 0.3–0.6), reflecting limited generalization under MAGI-1. \textbf{HiFi-Net} performs worst overall, with low T1 values ($<$ 0.39) and near-zero T0.1 on most splits, suggesting that its learned features fail to transfer to this commercial generator.

These results highlight two key trends. First, detectors with strong multi-generator training performance (TALL, DF-Adapter) transfer well to unseen commercial generators, achieving excellent or near-excellent robustness. Second, detectors such as CADDM and HiFi-Net, which are more fragile under distribution shifts, fail to generalize effectively despite maintaining higher AUC values. This reinforces the broader insight from our benchmark: aggregate metrics such as AUC can obscure vulnerabilities, while threshold-based measures (T1, T0.1) reveal substantial gaps in robustness to emerging generators.

\begin{table*}[t]
\centering
\caption{Supplementary results on MAGI-1 commercial generator.}
\label{tab:mag1_t1}

\begin{subtable}{\textwidth}
\centering
\adjustbox{max width=\textwidth}{%
\begin{tabular}{l ccc ccc ccc ccc ccc ccc}
\toprule
& \multicolumn{3}{c}{AniAudio~\cite{wei2024aniportrait}}
& \multicolumn{3}{c}{AniVideo~\cite{wei2024aniportrait}}
& \multicolumn{3}{c}{Hallo~\cite{xu2024hallo}}
& \multicolumn{3}{c}{Hallo2~\cite{cui2024hallo2}}
& \multicolumn{3}{c}{EMOPortraits~\cite{drobyshev2024emoportraits}}
& \multicolumn{3}{c}{LivePortrait~\cite{guo2024liveportrait}} \\
\cmidrule(lr){2-4}\cmidrule(lr){5-7}\cmidrule(lr){8-10}\cmidrule(lr){11-13}\cmidrule(lr){14-16}\cmidrule(lr){17-19}
Detector & AUC & T1 & T0.1
         & AUC & T1 & T0.1
         & AUC & T1 & T0.1
         & AUC & T1 & T0.1
         & AUC & T1 & T0.1
         & AUC & T1 & T0.1 \\
\midrule
CADDM~\cite{dong2023implicit}     & 0.88 & 0.29 & 0.29 & 0.95 & 0.39 & 0.39 & 0.91 & 0.53 & 0.53 & 0.96 & 0.62 & 0.62 & 0.97 & 0.62 & 0.62 & 0.97 & 0.64 & 0.64 \\
HiFi-Net~\cite{hifi_net_xiaoguo}  & 0.66 & 0.02 & 0.00 & 0.89 & 0.39 & 0.23 & 0.85 & 0.04 & 0.02 & 0.73 & 0.02 & 0.00 & 0.86 & 0.06 & 0.04 & 0.87 & 0.09 & 0.02 \\
TALL~\cite{xu2023tall}            & 1.00 & 1.00 & 1.00 & 1.00 & 1.00 & 1.00 & 1.00 & 1.00 & 1.00 & 1.00 & 1.00 & 1.00 & 1.00 & 1.00 & 1.00 & 1.00 & 1.00 & 1.00 \\
DF-Adapter~\cite{shao2025deepfake}& 1.00 & 1.00 & 1.00 & 1.00 & 0.99 & 0.99 & 1.00 & 0.99 & 0.99 & 1.00 & 0.99 & 0.99 & 1.00 & 1.00 & 1.00 & 1.00 & 1.00 & 1.00 \\
\bottomrule
\end{tabular}}
\end{subtable}
\end{table*}

\subsection{Hallo3 Academic Generator Results}
\label{suppl-subsec:hallo3_results}

We evaluate detector robustness on Hallo3, a recent academic talking-head generator. Results are shown in Tab.~\ref{tab:hallo3_t1}. As with MAGI-1, we restrict evaluation to detectors operating on visual input.

\textbf{TALL} achieves perfect scores across all metrics (AUC, T1, T0.1), demonstrating excellent generalization to Hallo3. \textbf{DF-Adapter} also performs nearly perfectly, with T1 and T0.1 values consistently at 0.99–1.00 across all splits. By contrast, \textbf{CADDM} shows only moderate performance, with AUC ranging from 0.67 to 0.97 and T1 values spanning 0.24–0.91 depending on the split. \textbf{HiFi-Net} performs inconsistently: while it reaches very high scores on EMOPortraits (AUC = 1.00, T1 = 0.99), it collapses on other splits (e.g., T1 = 0.19 on AniAudio, T0.1 = 0.00 on LivePortrait), highlighting its fragility under generator and protocol shifts.

These findings mirror broader trends from our benchmark: detectors such as TALL and DF-Adapter maintain strong robustness on unseen generators, while CADDM and HiFi-Net struggle to generalize reliably. The variability in CADDM and the failure cases of HiFi-Net underscore the importance of evaluating with stricter thresholds, as aggregate metrics like AUC alone would obscure these weaknesses.

\begin{table*}[t]
\centering
\caption{Supplementary results on Hallo3 academic generator.}
\label{tab:hallo3_t1}

\begin{subtable}{\textwidth}
\centering
\adjustbox{max width=\textwidth}{%
\begin{tabular}{l ccc ccc ccc ccc ccc ccc}
\toprule
& \multicolumn{3}{c}{AniAudio~\cite{wei2024aniportrait}}
& \multicolumn{3}{c}{AniVideo~\cite{wei2024aniportrait}}
& \multicolumn{3}{c}{Hallo~\cite{xu2024hallo}}
& \multicolumn{3}{c}{Hallo2~\cite{cui2024hallo2}}
& \multicolumn{3}{c}{EMOPortraits~\cite{drobyshev2024emoportraits}}
& \multicolumn{3}{c}{LivePortrait~\cite{guo2024liveportrait}} \\
\cmidrule(lr){2-4}\cmidrule(lr){5-7}\cmidrule(lr){8-10}\cmidrule(lr){11-13}\cmidrule(lr){14-16}\cmidrule(lr){17-19}
Detector & AUC & T1 & T0.1
         & AUC & T1 & T0.1
         & AUC & T1 & T0.1
         & AUC & T1 & T0.1
         & AUC & T1 & T0.1
         & AUC & T1 & T0.1 \\
\midrule
CADDM~\cite{dong2023implicit}     & 0.67 & 0.24 & 0.24 & 0.70 & 0.40 & 0.40 & 0.93 & 0.76 & 0.76 & 0.82 & 0.64 & 0.64 & 0.97 & 0.91 & 0.91 & 0.93 & 0.69 & 0.69 \\
HiFi-Net~\cite{hifi_net_xiaoguo}  & 0.84 & 0.19 & 0.18 & 0.88 & 0.30 & 0.27 & 0.91 & 0.24 & 0.20 & 0.65 & 0.24 & 0.12 & 1.00 & 0.99 & 0.99 & 0.87 & 0.02 & 0.00 \\
TALL~\cite{xu2023tall}            & 1.00 & 1.00 & 1.00 & 1.00 & 1.00 & 1.00 & 1.00 & 1.00 & 1.00 & 1.00 & 1.00 & 1.00 & 1.00 & 1.00 & 1.00 & 1.00 & 1.00 & 1.00 \\
DF-Adapter~\cite{shao2025deepfake}& 1.00 & 0.99 & 0.99 & 1.00 & 1.00 & 1.00 & 1.00 & 1.00 & 1.00 & 1.00 & 0.99 & 0.99 & 1.00 & 1.00 & 1.00 & 1.00 & 0.99 & 0.99 \\
\bottomrule
\end{tabular}}
\end{subtable}
\end{table*}

\section{Additional Explainability Results}
\label{suppl-sec:additional_gradcam}
The main paper (Section 4.4) provides Grad-CAM visualizations for the TALL detector. This section includes additional Grad-CAM results for TALL and DeepFake-Adapter, the best two benchmarked detectors. These visualizations illustrate the image regions these models focus on, offering insights into their decision-making and potential biases across different generators and protocols.

\subsection{Grad-CAM Visualizations for TALL~\texorpdfstring{\cite{xu2023tall}}{[\textbf{drobyshev2024emoportraits}]}}
\begin{figure}[h!]
    \centering
    \includegraphics[width=\linewidth]{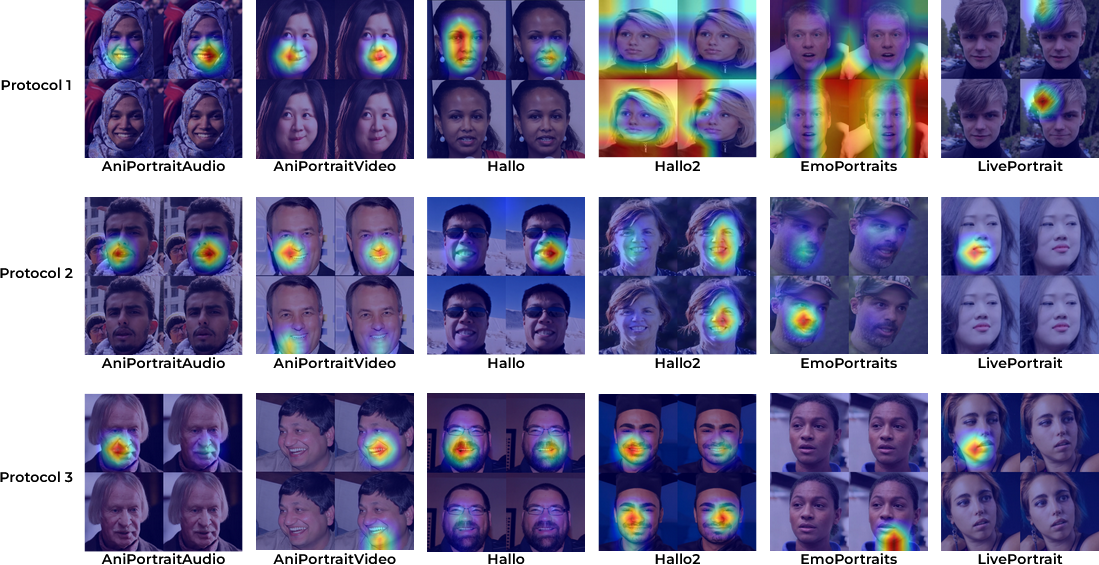}
    \caption{Grad-CAM visualizations of the TALL detector on correctly classified, high-confidence fake samples across Protocols 1–3 (one sample per generator). TALL predominantly focuses on facial features such as the eyes, nose, and mouth, but strategically shifts its attention to background or peripheral regions for certain generators. In Protocol 1, where the model has seen all generators during training, it focuses on background cues for Hallo2 and EMOPortraits, suggesting that these regions carry distinctive generative artifacts. In Protocol 3, the detector localizes to the neck region for EMOPortraits, while maintaining facial focus elsewhere. These examples reflect TALL’s adaptive attention and its ability to leverage generator-specific cues that contribute to its superior generalization and detection performance.}
    \label{fig:gradcam_tall}
\end{figure}

Figure~\ref{fig:gradcam_tall} presents Grad-CAM visualizations of the TALL detector across Protocols 1–3, using one correctly classified high-confidence fake sample per generator per protocol. These examples reflect cases where TALL confidently and accurately identifies manipulations, allowing us to interpret what visual evidence it relies on for detection.

Across most generators and protocols, TALL focuses its attention primarily on the facial region—including key semantic features such as the eyes, nose, and mouth—indicating that it has learned to localize and exploit common deepfake artifacts. Notably, in Protocol 1, where the model is trained on all generators (i.e., no generator shift), TALL relies exclusively on background regions for Hallo2 and EMOPortraits. This behavior contrasts with its facial focus for other generators and suggests that TALL has learned to exploit generator-specific cues in the background—clues that other detectors likely overlook.

In Protocol 3, where both identity and generator differ between train and test, TALL continues to rely on consistent facial features for most generators. However, it attends to the neck region for EMOPortraits, which may again indicate an adaptation to persistent artifacts in that region. Importantly, all examples shown here are correct classifications, underscoring that TALL’s attention—whether focused on facial or peripheral features—is meaningfully aligned with its high performance.

These visualizations suggest that TALL does not overfit to a single detection heuristic but instead generalizes flexibly across generators by identifying the most informative regions—be they facial or contextual—for each case. This ability to adaptively shift focus may contribute to its strong generalization under distribution shifts.



\subsection{Grad-CAM Visualizations for DeepFake-Adapter~\texorpdfstring{\cite{shao2025deepfake}}{[\textbf{shao2025deepfake}]}}
\begin{figure}[h!]
    \centering
    \includegraphics[width=\linewidth]{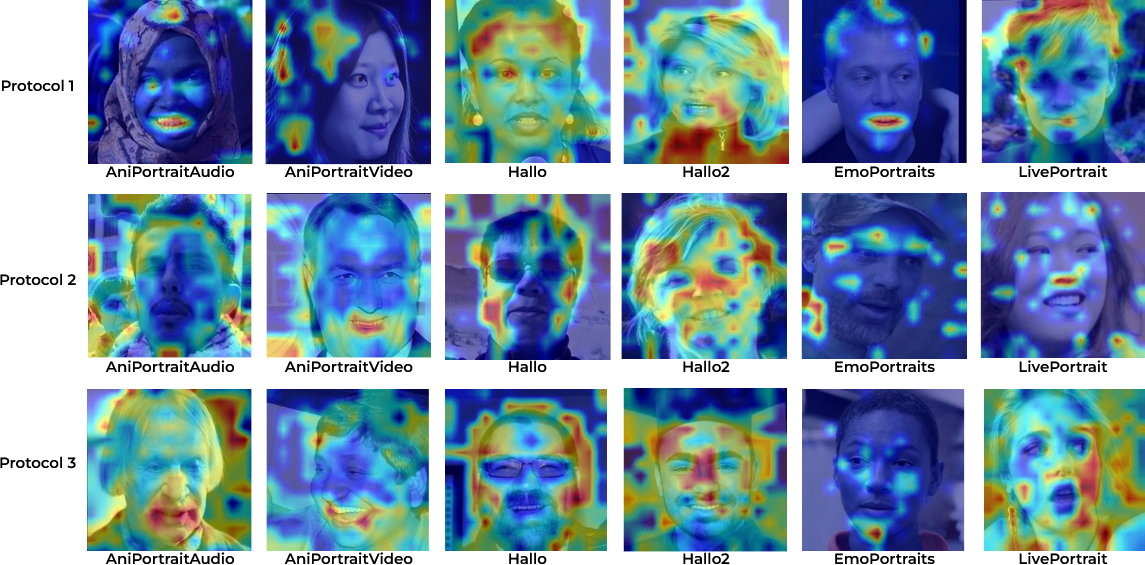}
    \caption{Grad-CAM visualizations of DeepFake-Adapter across Protocols 1–3 (same samples as used for TALL). Compared to TALL, DF-Adapter shows broader and less localized attention maps, often covering both facial and background regions with reduced spatial focus. While it identifies meaningful areas in some cases (e.g., the mouth region in EMOPortraits under Protocol 1), attention weakens in Protocols 2 and 3, especially for EMOPortraits, correlating with its performance drop. The model appears to rely on a mixture of weak signals across the image rather than generator-specific artifact patterns, reflecting its less consistent generalization under distribution shifts.}
    \label{fig:gradcam_dfadapter}
\end{figure}

Figure~\ref{fig:gradcam_dfadapter} presents Grad-CAM results for the DeepFake-Adapter model across Protocols 1–3 using the same sample images previously analyzed for TALL. Compared to TALL, whose attention maps are generally well-localized around facial or generator-specific regions, DF-Adapter displays broader and more diffuse attention, often spanning both the face and background without a clearly focused region of interest.

In many cases, DF-Adapter does attend to relevant facial areas—particularly the mouth and eyes—but the heatmaps suggest less spatial precision. This pattern is observed across most generators and protocols, indicating that DF-Adapter may rely on a combination of weak signals distributed across the image rather than distinct artifact cues. For example, in Protocol 1, the attention map for EMOPortraits is focused around the mouth, a region commonly manipulated in talking-head deepfakes. However, in Protocols 2 and 3, the attention on EMOPortraits degrades significantly, with minimal relevant focused regions, potentially explaining the model’s poor generalization to this generator under unseen conditions.

The results also highlight DF-Adapter’s less adaptive generalization behavior: while it sometimes leverages background information (as TALL does for Hallo2 and EMOPortraits), it lacks the targeted selectivity shown by TALL. This likely contributes to its performance drop under Protocols 2 and 3, where subtle artifacts become harder to detect without robust, focused visual strategies.

\section{Demographic Bias Audit} 

To assess the fairness of the benchmarked detectors and understand biases inherited from the source datasets (FFHQ, CelebV-HQ), we conducted a demographic audit on our test set using classifications from the FairFace model~\cite{karkkainen2021fairface}. We analyzed detector performance across three axes: race, gender, and age. Tab~\ref{tab:supp_demographic} summarized below indicate minimal bias along gender lines but reveal notable performance disparities across racial and age subgroups. This analysis highlights an important area for future work in both dataset creation and algorithmic fairness for deepfake detection.
\begin{table}[!t]
\centering
\small
\caption{\textbf{Detector performance by demographics.} Disparities appear across race and age, while gender differences are minimal.}
\label{tab:supp_demographic}

\begin{subtable}[t]{0.46\textwidth}
\centering
\caption*{\textbf{(a) Race}}
\begin{tabular}{lc}
\toprule
Group & Accuracy \\
\midrule
Asian            & 88.89\% \\
African American & 81.58\% \\
Indian           & 87.23\% \\
White            & 87.55\% \\
\bottomrule
\end{tabular}
\end{subtable}
\hfill
\begin{subtable}[t]{0.46\textwidth}
\centering
\caption*{\textbf{(b) Gender}}
\begin{tabular}{lc}
\toprule
Group & Accuracy \\
\midrule
Female & 87.77\% \\
Male   & 86.98\% \\
\bottomrule
\end{tabular}
\end{subtable}

\vspace{2pt}

\begin{subtable}[t]{\columnwidth}
\centering
\caption*{\textbf{(c) Age Group}}
\begin{tabular}{lc}
\toprule
Age Group & Accuracy \\
\midrule
0--2     & 95.00\% \\
3--9     & 93.88\% \\
10--19   & 90.91\% \\
20--29   & 87.90\% \\
30--39   & 84.48\% \\
40--49   & 82.56\% \\
50--59   & 85.37\% \\
60--69   & 94.12\% \\
70+      & 80.00\% \\
\bottomrule
\end{tabular}
\end{subtable}
\end{table}

\clearpage

\end{document}